\documentclass{article}

\usepackage{PRIMEarxiv}

\usepackage[utf8]{inputenc} 
\usepackage[T1]{fontenc}    
\usepackage{hyperref}       
\usepackage{url}            
\usepackage{booktabs}       
\usepackage{amsfonts}       
\usepackage{nicefrac}       
\usepackage{microtype}      
\usepackage{amsthm}

\usepackage{lipsum}
\usepackage{fancyhdr}       
\usepackage{graphicx}       
\usepackage{float}
\graphicspath{{media/}}     
\usepackage{subcaption}
\usepackage{amsmath}
\usepackage{tikz}
\usepackage[linesnumbered,ruled,vlined]{algorithm2e}
\usepackage[utf8]{inputenc}
\newtheorem{theorem}{Theorem}[section]

\newtheorem{assumption}{Assumption}[section]

\newtheorem{definition}[theorem]{Definition}

\newtheorem{remark}{Remark}

\title{A machine learning approach for image classification in synthetic aperture RADAR
}

\author{
 Romina Gaburro \\
  Department of Mathematics and Statistics\\
  University of Limerick\\
  Health Research Institute (HRI) \\
  \texttt{romina.gaburro@ul.ie} \\
   \And
 Patrick Healy \\
  Department of  Computer Science and Information Systems\\
  University of Limerick\\
  \texttt{patrick.healy@ul.ie} \\
  \And
 Shraddha Naidu \\
 Department of Mathematics and Statistics\\
  University of Limerick\\
  \texttt{shraddha.naidu@ul.ie} \\
  \And
Clifford Nolan \\
  Department of Mathematics and Statistics\\
  University of Limerick\\
  \texttt{clifford.nolan@ul.ie} \\
}

\begin{document}
\maketitle
\numberwithin{equation}{section}

\begin{abstract}
We consider the problem in Synthetic Aperture RADAR (SAR) of identifying and classifying objects located on the ground by means of Convolutional Neural Networks (CNNs).  Specifically, we adopt a single scattering approximation to classify the shape of the object using both simulated SAR data and reconstructed images from this data, and we compare the success of these approaches. We then identify ice types in real SAR imagery from the satellite Sentinel-1. In both experiments we achieve a promising high classification accuracy ($\geq$75\%). Our results demonstrate the effectiveness of CNNs in using SAR data for both geometric and environmental classification tasks. Our investigation also explores the effect of SAR data acquisition at different antenna heights on our ability to classify objects successfully.
\end{abstract}


\section{Introduction}
Synthetic Aperture RADAR (SAR) is a RADAR-based imaging technology, distinguished by its ability to generate detailed images of regions of the Earth's surface or objects thereon, irrespective of the time of day or prevailing weather conditions. SAR imaging involves a moving antenna, typically mounted on an aircraft or satellite, which traverses a flight path while emitting pulses of electromagnetic radiation. These pulses interact with the terrain being imaged and scatter back towards the antenna, where the received signals are processed to reconstruct an image of the terrain \cite{SAR-Review, nolan-cheney}. 

The received data depends on two variables, namely the fast time $t$, measuring the time of flight of the radio wave and also the slow time variable $s$, which parametrises the position of the antenna along its flight track. The forward SAR operator maps the so-called \textit{target reflectivity function} $V$ to the scattered waves received by the antenna as in \eqref{eq:data_eq}. The target reflectivity function $V$ is responsible for the backscattered RADAR signal, and it is also the object that we normally wish to reconstruct or image. The SAR inverse problem aims in fact to reconstruct $V$ to form an image of the terrain from the received RADAR signals.

The accurate detection of objects by SAR data holds paramount importance across a wide spectrum of applications, including detailed Earth observation and environmental monitoring, precise biomass estimation for ecological studies and resource management, and the critical monitoring of glacial dynamics and ice sheet changes \cite{T-R-P-D}. Furthermore, the unique capabilities of SAR imagery, particularly its all-weather and day-night operational capacity, render it invaluable for military applications, notably in the surveillance and detection of maritime vessels \cite{C-A-C-W-H-L}, contributing significantly to maritime security and strategic intelligence. 

The characteristics of speckle noise, arising from the nature of the RADAR signal and the interaction with multiple scattering elements within the backprojected SAR image, require specialised processing techniques. Moreover, the geometric distortions inherent in SAR acquisitions, such as foreshortening, layover, and shadow, further compound the challenges in accurate object interpretation and detection \cite{SAR-Noise}. The development of robust object detection algorithms for SAR imagery often involves sophisticated noise reduction techniques, feature extraction methods tailored to SAR data characteristics (e.g., polarimetric information, texture analysis), and advanced Machine Learning (ML) approaches to effectively discern objects of interest from the complex background and inherent artifacts \cite{SAR-ARtifcat}. Note that in the context of this study, an artifact refers to a feature, pattern, or distortion in the SAR image that does not represent the true physical characteristics of the observed scene.

In this work, we combine classical SAR mathematical methodologies with Convolutional Neural Networks (CNNs) to detect and classify objects located on the surface of the Earth. We focus on four distinct scenarios: $i)$ shape detection of a single scatterer; $ii)$ multi-scatterer distinction; $iii)$ circular scatterer radius detection and number of scatterer detection, drawing a comparison with Electrical Impedance Tomography (EIT) \cite{C,G-H-N-N}; $iv)$ ice type detection.

 Our focus is on classifying the target reflectivity function $V$ introduced above from both the raw RADAR data and from the reconstructed SAR image itself (we will refer to this latter as \textit{backprojected SAR image data} or simply as \textit{SAR image data}). To this end, we train our ML algorithms to classify $V$ using both raw data and reconstructed images from such data.  In particular, we are interested in how successful the classification may be from just the raw SAR data itself, compared to the more computationally expensive training data obtained from image reconstructions, based on a standard backprojection procedure (see Algorithm \ref{alg:backprojection} below). We refer to \cite{shape,shape_deep} for further reading on shape detection and classification with SAR data.

As mentioned above, a persistent challenge in SAR imagery is the inherent presence of speckle noise, a multiplicative granular pattern that degrades image quality, resulting in grainy and often blurred visual representations of the terrain being imaged \cite{BrettBorden_1997}.  This noise significantly hinders the accurate identification of objects captured by SAR satellites, including prominent missions such as Sentinel-1 \cite{data,Sentinel1}. The complexities and diverse methodologies used in object detection within SAR images are reviewed in \cite{ZHERDEV-M-P-F, Y-W-Z-P-Z-Z, G-S-S-P-W}. Here, we demonstrate that the common pitfalls of SAR imagery - such as speckle noise, image blurriness, and indistinct edges \cite{SAR-Noise} - can be effectively overcome using ML techniques, specifically CNNs and SAR data (both raw and backprojected SAR image data). 

Starting with the first scenario $i)$ of shape detection, we recall that this problem is pertinent to applications including non-destructive testing and indoor environment analysis. In \cite{B-W-G-K}, shapes are detected using SAR image data based on a circularity parameter $\eta_c$, which is a function of the object's perimeter and the number of pixels it covers in a terrain. However, this method requires \textit{a-priori} information on these parameters and $\eta_c$ to regularise the SAR inverse problem. Here we propose using CNNs (Section \ref{sec:cnn}), which do not require any \textit{a-priori} information. 

SAR image data has been used in the past to classify shapes of tundra lakes using deep learning \cite{D-S-K-A-L-M} and for ship/object detection \cite{L-X-S-G-W}. In \cite{K-T-B-B}, the classification of SAR data (based on the material of the object being imaged) was done with raw SAR data, instead of backprojected SAR image data. It is shown in \cite{K-T-B-B} that the raw SAR data outperforms backprojected SAR images in the classification of materials. The image reconstruction method adopted in \cite{K-T-B-B} was the so-called Omega-K algorithm \cite{Omega-K}. Briefly, the Omega-K algorithm performs a Fourier transform on the SAR data, applies matched filtering in both azimuth and range, conducts a Stolt mapping, and then performs an inverse Fourier transform to generate the image. 

Here, we test whether this outperformance of the raw SAR data over backprojected SAR image data is also true for shape detection. To test if raw SAR data outperforms SAR image data for shape classification, we first use a CNN to classify the target reflectivity function $V$ using raw SAR data, and then compare this with the results of classification using CNNs on backprojected SAR image data (Section \ref{sec:shape_results}). We show that raw SAR data offers better or identical classification (90.00-100\% accuracy) than backprojected SAR image data for shape detection (80.00-99.00\%), confirming the results in \cite{K-T-B-B} and extending them also to the task of shape detection addressed in this paper. 

We detect the shape of an object limited to a planar surface (Section \ref{sec:shape_results}). The specific shapes investigated here are detailed in Table \ref{tb:shape_types}. To simulate our SAR data, we employ a single scattering model (Section \ref{sec:single_scatter}). To collect SAR data in our simulations, we use an antenna traversing a circular flight path \eqref{eq:flight_track_x(S)} at a fixed hight $h$. We collect SAR data from the antenna at three heights $h$ (in meters), specifically $h = 0,5,10$. It will be understood throughout the entire paper that all lengths $h$ are expressed in meters and that, to ease the notation, we will simply denote them by means of their corresponding value $h$. We use simple backprojection (Algorithm \ref{alg:backprojection}) for the reconstructed images as this is a proof-of-concept paper, and we want to start with the simplest image formation algorithm. As we show that raw SAR data offers superior classification over backprojected SAR image data (Section \ref{sec:shape_results}) in scenario $i)$, we use raw SAR data gathered by Sentinel-1 only for the simulations in scenarios $ii)$ and $iii)$ (Sections \ref{sec:multi_Scatter} and \ref{sec:EITComp}, respectively). We use SAR images data only for scenario $iv)$ (Section \ref{sec:ice_results}), where raw data are not available.

Scenario $ii)$ of multi-scatterer distinction (Section \ref{sec:multi_Scatter}) uses raw SAR data only, collected from an antenna flying along a circular flight path \eqref{eq:flight_track_x(S)} at the fixed height $h=5$. This choice is dictated by the results of Section \ref{sec:shape_results} where the setup of collecting SAR raw data at $h=5$ offers the highest classification accuracy (100.00\%). We consider the simple case of multiple circular bump scatterers \eqref{eq:circular_bump}, where the task is to determine the maximum radii scatterers are allowed to have in order for our CNN to classify them as distinct objects. As expected from an intuitive point of view, if the two bumps occupy a majority of the region of interest, achieving $100\%$ accuracy in multi-scatter detection might not be feasible, although the performance level overall remains quite good to allow to distinguish the main features of the circular bumps also for large enough values of the radii of the two bumps (>90.00\% accuracy overall).

The case $h=0$ considered in scenario $i)$ is particularly significant, as it serves as a direct comparison to ground-based SAR. Our findings show that classification on raw SAR data is either identical or superior to classification on the backprojected SAR images collected, demonstrating that raw SAR data often retains more useful information than backprojected SAR image data. 

In scenario $iii)$ the detection of scatterer radii is addressed for the case of raw SAR data collected from an antenna flying along the circular path \eqref{eq:flight_track_x(S)} at a fixed height $h=0$ only as this case allows us to compare our results with those obtained in \cite{G-H-N-N} via Electrical Impedance Tomography (EIT), where a static imaging modality based on the reconstruction of the conductivity of materials by electrostatic measurements is performed on its surface \cite{Bo,C,U}. To compare the performance of classifying SAR data via CNNs with the results obtained in \cite{G-H-N-N}, we test two of the scenarios considered in \cite{G-H-N-N} -- that of the detection of the number of (possibly) multiple inclusions within an object and that of a circular inclusion radius detection (Section \ref{sec:EITComp}). We show that for the inclusion radii detection, when combined with neural networks, EIT outperforms SAR, but in the case of determining the number of inclusions (scatterers), SAR outperforms EIT significantly. 

In the final scenario $iv)$ of ice type detection, we utilise the Sentinel-1 dataset \cite{data} of the Belgica Bank in Greenland \cite{Boulze-Korosov-Brajard_2020}. SAR images from Sentinel-1 have been successfully used for object classification in urban environments \cite{kumar2021urban}. The dataset \cite{data} includes SAR backprojected images and the corresponding ice type labels provided by geological experts. The ice types considered here are listed in Table \ref{tb:Ice_Type}. 

 The task of classifying ice types is important as the melting of ice glaciers is an important indicator of climate change and sea-level variations \cite{Mi-Ra,von2006parameterization,T-R-P-D}. This process is currently labour-intensive, demanding many hours from expert analysts. The difficulty in ice-type classification with SAR images arises as different sea ice types can produce similar SAR responses \cite{Ice-Type-Class}. Additionally, a key indicator of ice type is the thickness of ice cover, but as this thickness cannot be determined by SAR images alone \cite{von2006parameterization} -- the classification of ice types poses a challenge. Previously, sea ice has been classified using information such as greyness level and texture \cite{ice95}. CNNs have shown success in classifying ice types (four categories) with SAR images \cite{Boulze-Korosov-Brajard_2020} - here we address the challenge of the more intricate classification of ice into the 8 types listed in Table \ref{tb:Ice_Type}, extending the results of \cite{Boulze-Korosov-Brajard_2020} to include Mountain ice, Water group, Glaciers and Icebergs. We implement a classifier by utilising the comprehensive dataset provided by \cite{data}. Using the CNN introduced in Section \ref{sec:cnn}, we achieve an accuracy of 75.50\% in detecting different ice types (Section \ref{sec:ice_results}). 

The paper is organised as follows: in Section \ref{sec:single_scatter} we describe the single-scattering forward model used to simulate the SAR data. Section \ref{sec:cnn} is dedicated to the description of the CNN used for our classifications. In Section \ref{sec:data_gen}, we describe the process of simulating the raw scattered data, $data(t,s)$, and the process of generating SAR image data using backprojection (Algorithm \ref{alg:backprojection}) along with the terrains simulated for our classification tasks. Section \ref{sec:results} contains the results of this paper. In Subsection \ref{sec:shape_results} we present the results of the shape detection task. In Subsection \ref{sec:multi_Scatter}, we detect the presence of multiple scatterers with varying radii, where the resolution limitations of our algorithms plays an important role. As an example, we identify the largest scatterer radius that can be reliably detected with 100.00\% accuracy. In Subsection \ref{sec:EITComp}, we compare the performance of SAR with the results obtained via EIT in \cite{G-H-N-N}, and in Subsection \ref{sec:ice_results}, we present the results for ice type detection. Finally, we conclude our findings in Section \ref{sec:conclusion}.

\section{SAR model and backprojection}

\label{sec:single_scatter}

Our model for each component of the electromagnetic field is based on the scalar wave equation as follows: 
\begin{equation}
    \left( \nabla^2 - \frac{1}{c^2(\mathbf{x})}\partial_t^2\right)u(t,\mathbf{x})=f(\mathbf x,t),
    \label{eq:wave_eq}
\end{equation}
where $u$ is a component of the electromagnetic wave vector field. Here, $\mathbf{x}\in \mathbb{R}^3$ denotes the position vector, $t \in \mathbb{R}_+$ is the (fast) time, where $\mathbb{R}_{+} = \left\{t\in\mathbb{R}\: :\: t>0\right\}$, and $c=c(\mathbf{x})$ is the wave propagation speed. \eqref{eq:wave_eq} is a commonly used approximation for the wave field components, although the correct model for wave propagation is Maxwell's equations \cite{nolan-cheney,ch,ch-bo,lopez2011inverse}. %
Here we consider a wave being transmitted from a source $f(\mathbf x,t)$, reflected by an object in a certain region of interest (ROI) $\Omega\subset\mathbb{R}^3$ (often called the \textit{scene}), located on the terrain. This reflected wave may then return to the receiver, which we assume to coincide with the source as per Assumption \ref{assumption monostatic} below.

\begin{assumption}\label{assumption monostatic}
   We assume a monostatic SAR system, i.e., the source and receiver are coincident. 
   \end{assumption}
    \begin{assumption}\label{assumption Born}
    We also assume waves scatter from the ROI $\Omega$ just once before returning to the transceiver, and in particular, we will assume that the so-called \textit{Born approximation} is valid in Equation (\ref{eq:wave_ref}) \cite{lopez2011inverse}.
\end{assumption}
\begin{assumption}\label{assumption speed constant}
    We assume a constant speed of wave propagation in the intervening region between the transceiver and the ROI $\Omega$, i.e.,  $c(\mathbf{x})=c_0=1.$
\end{assumption}
\noindent Under Assumption \ref{assumption Born}, the total scattered field $u^{sc}$ is approximated as the solution of the following linearised wave equation:
\begin{equation}
    \left( \nabla^2 - \frac{1}{c_0}\partial_t^2 \right)  u^{sc}(\mathbf{x},t) =- V(\mathbf{x})\partial_t^2u_0(\mathbf{x},t).
   \label{eq:wave_ref}
\end{equation}
We write the total field $u$ as
\begin{equation}
    u=u_0+u^{sc},
    \end{equation}
where $u_0$ denotes the incident field, $u^{sc}$ has been introduced in \eqref{eq:wave_ref} and $V(\mathbf{x}) = \frac{1}{c_0^2} - \frac{1}{c^2(\mathbf{x})}$ is the \textit{target reflectivity function} of the scene we are imaging (see \cite{nolan-cheney} for a detailed explanation of the above linearised model). Let $G_0(t,\mathbf{x})$ be the Green's function for \eqref{eq:wave_ref}, which can be explicitly written as
\begin{equation}\label{Green's function}
G_0(t,\mathbf{x}) = \frac{\delta(t-|\mathbf{x}|/c_0)}{4\pi |\mathbf{x}|},
\end{equation}
where $\delta$ is the Dirac delta function (distribution). Thus, combining \eqref{eq:wave_ref} and \eqref{Green's function}, we have
\begin{equation}\label{u scattered}
    u^{sc}(t,\mathbf{x}) = \iint G_0(t-\tau,\mathbf{z} -\mathbf{x})V(\mathbf{z})\partial^2_{\tau}u_0(\tau,\mathbf{z})\text{ d}\tau \text{ d}\mathbf{z}.
\end{equation}
If the monostatic SAR system flies along a flight path 
\begin{equation}\label{flight path}
\gamma:[s_{min},s_{max}]\to\mathbb R^3,
\end{equation}
then substituting for $u_0$ in \eqref{u scattered}, we obtain the following idealised model for the scattered field $u^{sc}$ due to a point source $f(\mathbf x,t)=\delta(\mathbf x,t)$:
\begin{align}
u^{sc}(t,\gamma(s)) = \iint \frac{-\omega^2e^{-i\omega(t - 2|\mathbf{z} -\gamma(s)|/c_0)}}{(4\pi)^2 |\mathbf{z} -\mathbf{x}(s)|^2}\: V(\mathbf{z}) \text{ d}\omega \text{ d}\mathbf{z}, \label{eq:u_sc}
\end{align}
where $\omega$ is the temporal ($t$) frequency. \eqref{eq:u_sc} is an \emph{oscillatory integral} \cite{duistermaat.11} and it shows that we may model the scattered field using a {\em Fourier integral operator} \cite{duistermaat.11}, which maps the scene $V$ to the measured data $u^{sc}(t,\gamma(s))$, for $s\in [s_{min},s_{max}]$.  We omit the detailed steps in this simplification; for an in-depth formulation, we refer the reader to \cite{nolan-cheney}. It is also possible to consider a more refined model of scattered waves, which takes into account a more realistic source, including features like antenna beam patterns \cite{nolan-cheney}, but we do not do that here as it is not necessary for demonstrating our essential points about ML classification. 

One of our goals here is in fact to assess whether ML classification of SAR imagery performs better by training with the raw SAR data rather than the corresponding SAR backprojected image data. Furthermore, the factor $-\omega^2$ that arises in \eqref{eq:u_sc} is due to time-differentiation, and this aspect of the model is also not critical for our purpose here either. The geometrical spreading factor in the denominator of the integrand in \eqref{eq:u_sc} is also not important in this context. Thus, the essential aspects of our linearised scattering model are captured by the following simplified data model:
\begin{equation}
    data(t,s) = \int_{c_0t=2|\mathbf{z}-\gamma(s)|} V(\mathbf{z}) \text{ d}\mathbf{z}.
    \label{eq:circular integral}
\end{equation}

\begin{remark}
    Note that we have employed the oscillatory integral representation \cite{duistermaat.11} of the Dirac delta distribution to arrive at the formula in \eqref{eq:circular integral} from \eqref{eq:u_sc}.
\end{remark}

\begin{assumption}
    We make an idealised assumption that the ground coincides with $\mathbb{R}^2$ and that the reflectivity function $V(\mathbf{x})=V(x_1,x_2,x_3)$ has the form $V(x_1,x_2)\delta (x_3)$, so that we consider scatters that are concentrated at ground level, where $x_3=0$, i.e., $\Omega\subset\mathbb{R}^2$. This assumption is not crucial, but it eases the exposition of the main ideas in this work. 
\end{assumption}

\begin{figure}[H]
    \centering
\includegraphics[width=0.5\linewidth]{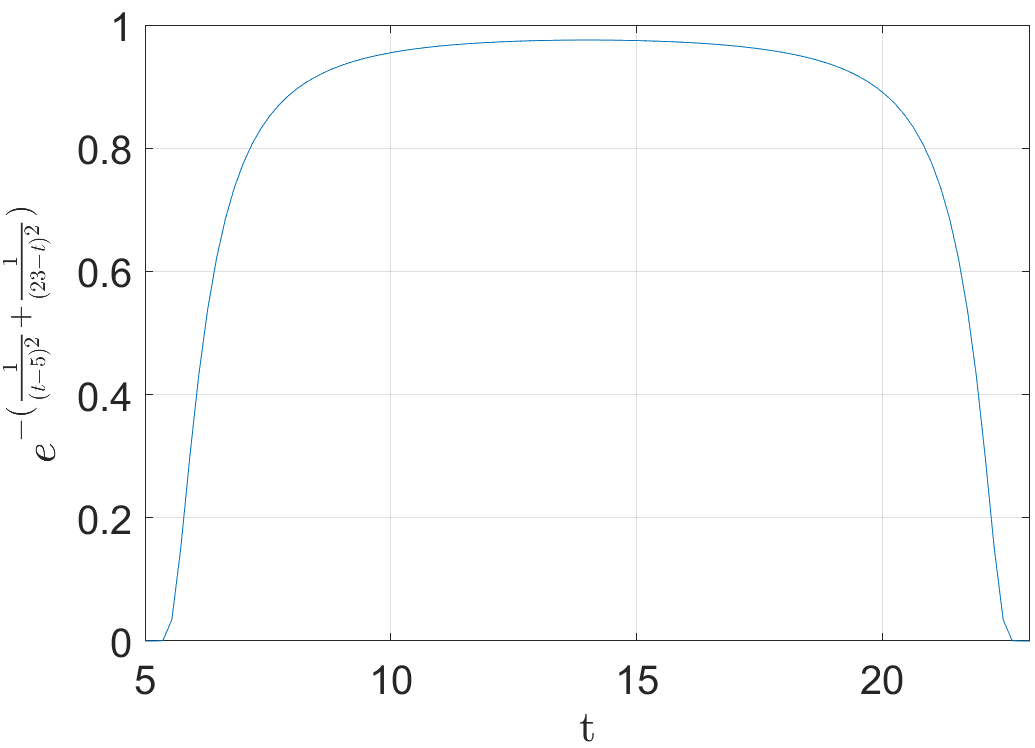}
    \caption{Smoothing function $\mu$ in the $t$ variable for raw SAR data (\eqref{eq:data_eq}).}
    \label{fig:mu}
\end{figure}

\begin{figure}[H]
    \centering
    \begin{subfigure}{0.32\textwidth} %
        \centering
        \includegraphics[width=\linewidth]{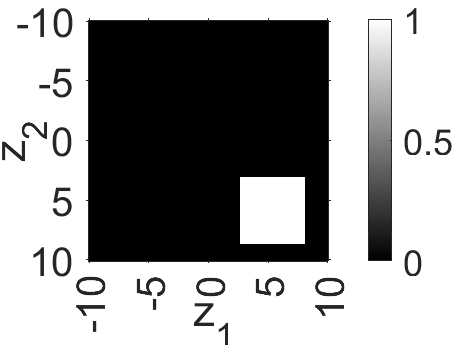} %
        \caption{}
        \label{fig:bs_1}
    \end{subfigure}
    \hfill 
    \begin{subfigure}{0.32\textwidth}
        \centering
        \includegraphics[width=\linewidth]{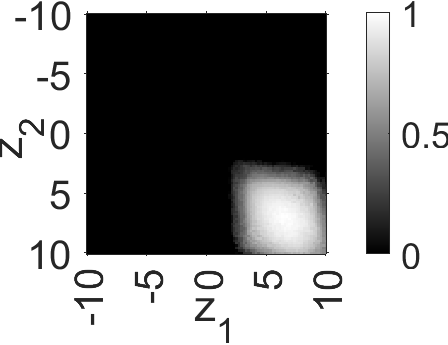} %
        \caption{}
        \label{fig:bs_2}
    \end{subfigure}
    \hfill
    \begin{subfigure}{0.32\textwidth}
        \centering
        \includegraphics[width=\linewidth]{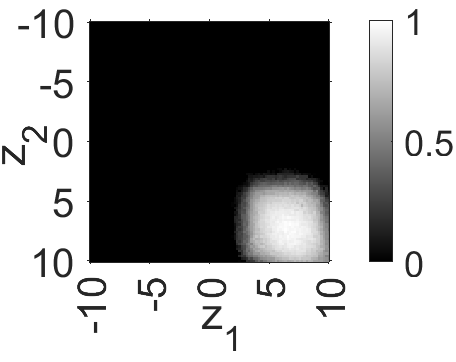} %
        \caption{}
        \label{fig:bs_3}
    \end{subfigure}
    \caption{Backprojected SAR images of the terrain $\Omega$ in \eqref{Omega} with a square-shaped bump of side length = 5.5 located in the lower right corner of the grid. The image is shown with and without smoothing. Figure \ref{fig:bs_1} shows the original terrain with the square bump. Figure \ref{fig:bs_2} shows the backprojected image of \ref{fig:bs_1} without smoothing and Figure \ref{fig:bs_3} shows the backprojected image of \ref{fig:bs_1} with the smoothing $\mu$ introduced in \eqref{eq:smooth}. Notice the protrusion in the top-left corner of the square bump in Figure \ref{fig:bs_2} has disappeared after the addition of smoothing $\mu$   \eqref{eq:smooth}.}
    \label{fig:back_smooth}
\end{figure}

\noindent Under this assumption, \eqref{eq:circular integral} becomes
\begin{equation}
    data(t,s) = \int_{c_0t=2|\mathbf{x}-\gamma(s)|} V(x_1,x_2) \text{ d}\mathbf{x}.
    \label{eq:data_eq}
\end{equation}
where we now have $\mathbf{x}=(x_1,x_2)\in\mathbb R^2$. This changes the region of integration in \eqref{eq:circular integral} from 3D to 2D. 
\begin{assumption} 
The antenna collecting the SAR data traverses a circular path (\eqref{eq:flight_track_x(S)}) at a fixed height $h$.
\end{assumption}
Thus, our data consists of a collection of integrals of the reflectivity function $V=V(\mathbf{x})$, taken over circles with centres at $(\gamma_1(s),\gamma_2(s))$ , with $s\in [s_{min}, s_{max}]$, and radii $\sqrt{(\frac{c_0t}{2})^2 -h^2}$. The SAR reconstruction problem is to recover $V$ from these circular integral measurements. Smoothing is applied to the raw SAR data (\eqref{eq:data_eq}) to reduce distortions and blurriness commonly present in SAR images (see Figure \ref{fig:back_smooth}). We add smoothing in the $t$ variable of the form, 
\begin{equation}
    \hat{D}(t,s) = data(t,s)\mu(t)=data(t,s) \cdot e^{-\left({(t-t_{min})^{-2}} +(t_{max}-t)^{-2}\right)}.
    \label{eq:smooth}
\end{equation}
Figure \ref{fig:mu} shows the behaviour of the smoothing function $\mu$ appearing in \eqref{eq:smooth}. The smoothed raw SAR data is processed to generate a terrain image using a technique called backprojection (see Algorithm \ref{alg:backprojection}). Backprojection gives an image of the scattering terrain, our ROI $\Omega\subset\mathbb{R}^2$, using the smoothed data $\hat{D}(t,s)$ as (see Assumption \ref{assumption Born}), 
\begin{equation}
    I(\mathbf{z}) = \int \delta \left( t - \frac{2|\mathbf{z}-\gamma(s)|}{c_0}
    \right)\hat{D}(t,s) \text{ d}s \text{ d}t,\qquad\textnormal{for\: any}\quad \mathbf{z}\in\Omega.\label{eq:SAR_Image}
\end{equation}
We refer to \cite{ch,ch-bo,ch-tomography,lopez2011inverse,M-N} for an in-depth reading on the backprojection algorithm (Algorithm \ref{alg:backprojection}).\\

Throughout the entire paper, all distances are expressed in meters.\\

For our simulations, we use a square grid scene as the terrain. The points in the grid correspond to $\mathbf{z}=(z_1,z_2,0)$, as all the scatters occur at $z_3=0$. To simplify our notation, we will simply denote these points by $\mathbf{z}=(z_1,z_2)\in\mathbb{R}^2$. We introduce our Region of Interest (ROI) $\Omega\subset\mathbb{R}^2$, 
\begin{equation}\label{Omega}
\Omega = \left\{\mathbf{z}=(z_1,z_2)\in\mathbb{R}^2\: : \: -10 \leq z_i \leq 10,\quad\textnormal{for}\quad i=1,2\right\}
\end{equation}
and discretise it into a 2D grid spanning $z_1,z_2 \in [-10,10]$ with 100 points in each direction. Although an abuse of notation, we will continue to call $\Omega$ its discretized version as just described.

\noindent Next, we describe the Convolutional Neural Networks used in this paper.

\section{Convolutional Neural Networks}
\label{sec:cnn}

A Convolutional Neural Network (CNN) is a neural network architecture that is commonly used for image classification. CNNs are a collection of \textit{layers} consisting of \textit{neurons} with \textit{weights} ($\mathbf{W}$) and \textit{biases} ($b$) associated with them. The weights and biases of a neural network are learnable parameters that are updated during training (see \eqref{eq:CNN:min}). Here, we use 7 layers in our CNN. For each layer $\ell \in \{1,..,7\}$ there is a function $f_{\ell}$ associated to $\ell$ which depends on the weights $\mathbf{W}^{(\ell)}$ and biases $b^{\ell}$ of $\ell$ (see \eqref{eq:f1}-\eqref{eq:f7}). Our input data is denoted by $\mathbf{p}^{(1)} \in \mathbb{R}^{P \times P \times C_{input}}$, with $P$ denoting the dimensions of the input data and $C_{input}$ denoting the number of \textit{channels} in the input data. A channel refers to a component of the input data that contains intensity values for a specific colour or aspect of the input data.
\begin{remark}\label{rm:chanels}
    Traditionally, a CNN processes images using 3 channels ($C_{input}=3$), each representing the red, green and blue (RGB) colour components. However, since the SAR images are monochromatic (grayscale), it will only require a single channel ($C_{input}=1$). Therefore, we omit the $\times C_{input}$ in our SAR analysis subsequently. The input data are simply $\mathbf{p}^{(1)} \in \mathbb{R}^{P \times P}$, where $P\times P$ is effectively the number of pixels in the image.
\end{remark}
\noindent Therefore, the CNN is represented by a composition function of type
\begin{equation}
\begin{aligned}
    \mathcal{F}:\mathbb{R}^{P\times P} \mapsto \mathbb{R},\\
    \mathcal{F}(\mathbf{p}^{(1)}) = \left(f_7 \circ f_6 \circ f_5 \circ f_4 \circ f_3 \circ f_2 \circ f_1\right)(\mathbf{p}^{(1)}),
    \end{aligned}
\label{eq:CNN_Covs}
\end{equation}
for any $\mathbf{p}^{(1)} \in \mathbb{R}^{P \times P}$. Here, we will set $P=100$ throughout the entire paper, except for the ice-type classification problem considered in Section \ref{sec:ice_results}, where we set $P=256$. The function $f_1$ is applied to $\mathbf{p}^{(1)}$ to obtain the collection of $K$ \textit{feature matrices} $\mathbf{p}^{(2)}$ via a convolutional layer, i.e.,
\begin{equation}\label{f1}
f_1: \mathbb{R}^{P \times P} \longrightarrow \mathbb{R}^{(P-K_f+1) \times (P-K_f+1)\times K}
\end{equation}
creates $K$ feature matrices each from a different filter (kernel), where $K_f$ is the filter size (kernel dimension). It is the presence of this layer that makes a neural network a CNN. We choose $K_f=13$ (see Remark \ref{rem:param_CNN}).  In particular,
\begin{equation}
         \left(f_1(\mathbf{p}^{(1)})\right)_{i,j,k} = \sum_{m=1}^{K_f}\sum_{n=1}^{K_f}\mathbf{p}^{(1)}_{i+m-1,j+n-1}\mathbf{W}^{(1)}_{m,n,k} + {b}^{(1)}_k  = (\mathbf{p}^{(2)})_{i,j,k},
         \label{eq:f1}
\end{equation}  
where $k \in\{1,...,K\}$ is the $k^{th}$ output feature matrix; the weights $\mathbf{W}^{(1)}_k\in\mathbb{R}^{K_f \times K_f}$ and $b^{(1)}_k \in \mathbb{R}$ are the weight matrix and bias associated with the $k^{th}$ filter respectively. The complete set of weights for this layer is $\mathbf{W}^{(1)} \in \mathbb{R}^{K_f \times K_f \times K}$. We use $K=1$ for all classifications except the ice-type detection (Section \ref{sec:ice_results}). As ice-type detection is more complex (more classes, larger images, more patterns to detect), we use $K=16$ in that case. 
In ML parlance, the convolution layer $f_1$ applies $K$ unique $K_f\times K_f$ filters to obtain, as output, the \textit{feature matrix} $\mathbf{p}^{(2)}_k$, for each feature filter $k \in\{1,\dots ,K\}$, as in \eqref{eq:f1}. The set of weights $\mathbf{W}^{(1)}$ is specifically trained to detect a particular pattern, or feature, within the input data $\mathbf{p}^{(1)}$. Consequently, the output of each filter, often referred to as a \textit{feature matrix} $\mathbf{p}^{(2)}_k$, represents the spatial distribution and intensity of the presence of that specific feature across the input. With $K$ filters, we are effectively learning to extract $K$ distinct types of features simultaneously. The filter size of $13\times 13$ is an arbitrary choice (see Remark \ref{rem:param_CNN}). 

The convolutional layer is followed by a \textit{batch normalisation layer} which is mathematically represented by a function 
\begin{equation}\label{f2}
f_2: \mathbb{R}^{M \times M \times K} \longrightarrow \mathbb{R}^{M \times M \times K},
\end{equation}
where $M = P-K_f -1$, is defined by
\begin{equation}\label{eq:f2}
\left(f_2(\mathbf{p}^{(2)})\right)_{i,j,k} = \gamma_k\left(\frac{(\mathbf{p}^{(2)})_{i,j,k} - \mu_k^{(2)}}{\sqrt{(\sigma_k^{(2)})^2 + \epsilon}}\right) + \beta_k = (\mathbf{p}^{(3)})_{i,j,k},
\end{equation}
and where $\mu_k^{(2)}, (\sigma_k^{(2)})^2 \in \mathbb{R}$ are the mean and variance of the $k$-th feature map $(\mathbf{p}^{(2)})_{k}$, $\gamma_k, \beta_k \in \mathbb{R}$  are learnable parameters (often called ``scale" and ``offset" respectively) associated with the $k^{th}$ feature matrix, and $\epsilon \in \mathbb{R}^+$ is the machine epsilon to prevent division by 0. 

This is followed by a \textit{Rectilinear unit (ReLu) layer} which is a function $f_3$,
\begin{equation}\label{f3}
f_3: \mathbb{R}^{M \times M \times K} \longrightarrow \mathbb{R}^{M \times M \times K},
 \end{equation}
 defined by
 \begin{equation}\label{eq:f3} 
   \left(f_3(\mathbf{p}^{(3)})\right)_{i,j,k} = \max\left(0, (\mathbf{p}^{(3)})_{i,j,k}\right) = (\mathbf{p}^{(4)})_{i,j,k},
 \end{equation}
which is followed by a \textit{max pooling layer}
\begin{equation}\label{f4}
f_4: \mathbb{R}^{M \times M \times K} \longrightarrow \mathbb{R}^{\frac{M}{2} \times \frac{M}{2} \times K},
\end{equation}
defined by
\begin{equation}\label{eq:f4}
   \left( f_4(\mathbf{p}^{(4)})\right)_{i,j,k} = \max_{0 \leq m,n < 2} \mathbf{p}^{(4)}_{2i+m,2j+n,k} = (\mathbf{p}^{(5)})_{i,j,k}. 
\end{equation}
This is followed by the \textit{fully connected layer} 
\begin{equation}\label{f5}
f_5: \mathbb{R}^{\frac{M}{2} \times \frac{M}{2} \times K} \longrightarrow \mathbb{R}^O,
\end{equation}
where $O$ is the number of output classes (e.g. 4 for shape detection, 8 for ice-type detection). The function $f_5$ is defined by 
\begin{equation}\label{eq:f5}
    f_5(\mathbf{\mathbf{p}^{(5)}}) = \mathbf{W}^{(5)} \cdot \text{vec}(\mathbf{p}^{(5)}) + b^{(5)} =  \mathbf{p}^{(6)},
\end{equation}
where $\mathbf{W}^{(5)} \in \mathbb{R}^{O \times  \frac{M^2}{4}\cdot K}$, is the weight matrix, $b^{(5)} \in \mathbb{R}^O$ is the bias, and 
\[\text{vec}: \mathbb{R}^{\frac{M}{2} \times \frac{M}{2} \times K } \longrightarrow \mathbb{R}^\frac{M^2 \cdot K}{4},\] is the operator that transforms a 3D matrix to a column vector. This is followed by the \textit{softmax layer}, 
\begin{equation}\label{f6}
f_6: \mathbb{R}^O \to \mathbb{R}^O,
\end{equation}
defined by the formula
 \begin{equation}
     \left(f_6(\mathbf{p}^{(6)})\right)_i = \frac{e^{\mathbf{p}^{(6)}_i}}{\sum_{j=1}^O e^{\mathbf{p}^{(6)}_j}} = \mathbf{p}^{(7)} \text{ for $i \in \{1, 2, \ldots, O\}$}.
     \label{eq:softmax}
 \end{equation}

\noindent The output $\mathbf{p}^{(7)}$ is a vector of probabilities, where each element $\mathbf{p}^{(7)}_i$ represents the predicted probability of the input data belonging to class $i$. 
This is followed by the final \textit{classification layer}
\begin{equation}\label{f7}
f_7: \mathbb{R}^O \to \{1, 2, \ldots, O\},
\end{equation}
which outputs the predicted final class of the image, defined as,
\begin{equation}
    f_7(\mathbf{p}^{(7)}) = \arg\max_i \mathbf{p}^{(7)}_i.
    \label{eq:f7}
\end{equation}
This layer simply selects the class with the highest predicted probability from the softmax output. In the case where $i$ is not unique for $\mathbf{p}^{(7)}$, one $i$ is randomly selected by the CNN algorithm as the output.
\begin{remark}
    The selection of hyperparameters, such as the filter size $K_f$ in the first convolutional layer $f_1$ in \eqref{eq:f1} and the pooling function in the fourth layer $f_4$ in \eqref{eq:f4} was initially made without explicit optimisation. This common practice in ML involves constructing a standard CNN with arbitrarily defined initial hyperparameters, subsequently adapting the network architecture based on performance metrics to better suit the specific dataset. Given the high classification accuracy achieved with this initial configuration, we have adopted the architectural structure given in \eqref{f1}-\eqref{eq:f7} throughout the remainder of this work. \label{rem:param_CNN}
\end{remark}
\noindent The specific dimensions per layer for the data used in this work are shown in Table \ref{tb:cnn}.
\begin{table}[H]
\centering
\begin{tabular}{c|c|c|c}
\textbf{Layer} & \textbf{Layer Name}       & \textbf{100x100 image} & \textbf{256x256 image} \\ \hline
$f_1$             & Convolutional Layer               & 88x88x$K$        & 244x244x$K$      \\
$f_2$             & Batch Normalization Layer & 88x88x$K$        & 244x244x$K$      \\
$f_3$             & ReLu Layer                & 88x88x$K$        & 244x244x$K$     \\
$f_4$             & Max Pooling Layer         & 44x44x$K$        & 122x122x$K$      \\
$f_5$             & Fully Connected Layer     & $O$          & $O$      \\
$f_6$             & Softmax Layer             & $O$                & $O$                \\
$f_7$             & Classification Layer      & 1                & 1               
\end{tabular}
\caption{Table showing the dimension changes for the inputs per layer of CNN. The $100 \times 100$ images are used for shape detection, multi-scatterer detection and scatterer radii detection (Section \ref{sec:shape_results}-\ref{sec:EITComp}). The $256 \times 256$ images are used for ice-type detection (Section \ref{sec:ice_results}).}
\label{tb:cnn}
\end{table}

\noindent The CNN learns by adjusting its internal parameters (weights and biases) to minimise a predefined loss function (\eqref{eq:cross}).  The overall function $\mathcal{F}$ in (\ref{eq:CNN_Covs}) depends on the collection of all learnable weights $\mathbf{W}^{(j)}$ and biases $b^{(j)}$ across all layers $j$ of the network. We denote the entire set of learnable parameters as $\Theta = \{\mathbf{W}^{(j)},b^{(j)}, \gamma_k, \beta_k\}$. The error in the prediction of the CNN can be written using the definition of $\mathcal{F}$ in \eqref{eq:CNN_Covs} and \eqref{eq:f1}-\eqref{eq:f7} as,
\begin{equation}
    \mathcal{L}(\Theta,\mathbf{p}^{(1)}) = g (\mathcal{F}(\Theta,\mathbf{p}^{(1)})).
\end{equation}

Here $\mathcal{L}$ is the loss function that quantifies the discrepancy between the CNN's predictions and the true classification, and $g$ is a non-negative valued function describing the error in prediction (\eqref{eq:cross}). Therefore, when we train the CNN, we aim to find optimal weights and biases $\Theta^*$ such that,
\begin{equation}
    \Theta^* =  \arg \min_{\Theta}\mathcal{L}(\Theta,\mathbf{p}^{(1)}). 
    \label{eq:CNN:min}
\end{equation}

\noindent The loss function $\mathcal{L} $ commonly used for classification problems is the cross-entropy loss function given by,
\begin{equation}
    \mathcal{L}(\Theta,\mathbf{p}^{(1)}) = -\frac{1}{N}\sum_{i=1}^N \sum_{o=1}^O y_{i,o}(\mathbf{p}^{(1)}) \log(\hat{y}_{i,o}(\Theta,\mathbf{p}^{(1)})),
    \label{eq:cross}
\end{equation}

where $N$ is the total number of samples in the training dataset, $O$ is the total number of output classes, $y_{i,o}$ is the binary indicator (1 if the $i^{th}$ input data $\mathbf{p}^{(1)}$ truly belongs to class $o$, and 0 otherwise) for the $i^{th}$ sample and $\hat{y_{i,o}}$ is the predicted probability of $i^{th}$ sample belonging to class $o$ as determined by the softmax function \eqref{eq:softmax}. $\mathcal{L}$ is higher for misclassifications and lower for correct classifications. Therefore, we aim to minimise $\mathcal{L}$ through training the CNN. 
 We solve \eqref{eq:CNN:min} using Adaptive Gradient with Momentum (ADAM) \cite{kingma-ba}. 
 \begin{definition}
     We define the accuracy of a CNN as the percentage of samples in the test dataset that were classified correctly. \label{def:acc}
 \end{definition}

 \begin{remark}
     To evaluate and compare the performance of our CNN we use a \textit{confusion matrix} (CM), a common tool for assessing classification accuracy. A CM is a square matrix $\mathbf{C}$ of size $n \times n$, where $n$ is the number of classes adopted in the classification task. Each entry $C_{ij}$ represents the number of times that samples from the actual class $i$ were predicted as class $j$. The rows of $\mathbf{C}$ correspond to the actual classes (the true class labels of the samples), where the columns correspond to the predicted classes (the labels assigned by the model). \label{rm:CM}
 \end{remark}
\section{Dataset generation}
\label{sec:data_gen}
This section presents the simulation and processing of the datasets we use as input to our CNNs introduced in Section \ref{sec:cnn}, \eqref{eq:CNN_Covs} - \eqref{f7}.


\subsection{Data Simulation for Shape Detection}
\label{sec:shape_sim}
We generate a terrain in the ROI $\Omega$ defined in \eqref{Omega} containing bumps of four different shapes that are embedded in a target reflectivity function introduced in \eqref{eq:data_eq}. These shapes are specifically a circular bump, a square bump, an elliptical bump and a rhombus-shaped bump. Specifically, we make the following assumption on the reflectivity function $V$ of $\Omega$.
\begin{assumption}
    The reflectivity function $V$ in $\Omega$ is characterised by a single localised bump (perturbation), which can take one of four shapes defined in Table \ref{tb:shape_types} below. 
\end{assumption}

\begin{table}[H]
\centering
\begin{tabular}{lc}
\textbf{Shape} & \textbf{Label O} \\ \hline
Circle         & 1                   \\
Square         & 2                   \\
Ellipse        & 3                   \\
Rhombus        & 4                  
\end{tabular}
\caption{The four types of bump shapes adopted in our simulated terrain for shape detection.}
\label{tb:shape_types}
\end{table}

\noindent These shapes of a circular bump, a square bump, an elliptical bump and a rhombus-shaped bump that lead to the reflectivity functions: $V_{circle}$ for circular bumps, $V_{square}$ for square bumps, $V_{ellipse}$ for elliptical bumps, and $V_{rhombus}$ for rhombus-shaped bumps. The centre for each shape $(z_1^0,z_2^0)$ is arbitrarily chosen to satisfy, 
\begin{equation}
    3  \leq z_1^0,z_2^0 \leq 6.
    \label{eq:centre}
\end{equation}

\begin{figure}[H]
    \centering

   \includegraphics[width=\textwidth]{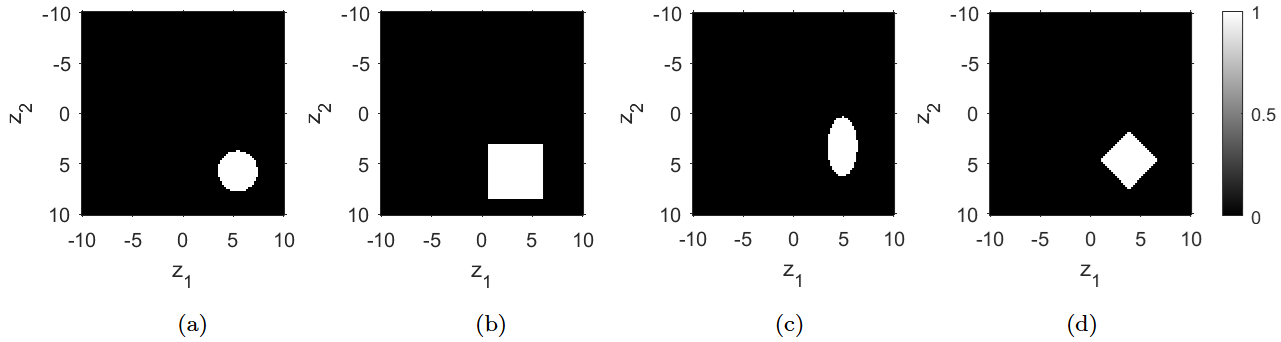}
    \caption{Terrains within the ROI $\Omega$ simulated for shape detection. Figures $3(a)$ - $3(d)$ show from left to right $V_{circle}$, $V_{square}$, $V_{ellipse}$ and $V_{rhombus}$.                                           The greyscale colorbar on the right shows the magnitude of the target reflexivity function $V$ in \eqref{eq:data_eq}.}
    \label{fig:diff_shapes} 
\end{figure}

\noindent We start by considering a circular bump centred at some point in $\Omega$. Let $C$ be the disk with centre $(z_1^0,z_2^0)$ and radius $r$,
\begin{equation}\label{disk}
C= \left\{ (z_1,z_2) \in \mathbb{R}^2: (z_1 -z_1^0)^2 + (z_2 - z_2^0)^2 \leq r^2\right\}. \notag \\ 
\end{equation}
Let $V$ be the reflectivity associated with $C$, defined by
\begin{equation}
V(z_1,z_2)_{circle} = \chi_{C}(z_1,z_2),
\label{eq:circular_bump}
\end{equation}
where $\chi_C(z_1,z_2)=1$ if $(z_1,z_2)\in C$ and equals zero otherwise, i.e., $\chi_D$ denotes the characteristic function of a set $D$. For the shape detection task, we use $r=2$ (Section \ref{sec:shape_results}). To define a square-shaped bump, we introduce,
\begin{align}
&S= \left\{ (z_1,z_2) \in \mathbb{R}^2: z_1^0 - \frac{s}{2}\leq z_1 \leq z_1^0 + \frac{s}{2}, z_2^0 - \frac{s}{2} \leq z_2 \leq z_2^0 + \frac{s}{2}\right\},\notag \\ 
    &V(z_1,z_2)_{square} =
\chi_{S}(z_1,z_2),
\label{eq:square_bump}
\end{align}
where $s=5.5$. To simulate an elliptical bump we introduce $V(z_1,z_2)_{ellipse}$,
\begin{align}
&E= \left\{ (z_1,z_2) \in \mathbb{R}^2: \frac{(z_1-z_1^0)^2}{a^2} +\frac{(z_2-z_2^0)^2}{b^2} \leq 1\right\},\notag \\ 
    &V(z_1,z_2)_{ellipse} =\chi_E(z_1,z_2),
\label{eq:ellipse_bump}
\end{align}
where $a=1.5$ is the minor and $b=3$ is the major axis of the ellipse. Finally, for the rhombus-shaped bump, we introduce $V(z_1,z_2)_{rhombus}$ by
\begin{align}
&R= \left\{ (z_1,z_2) \in \mathbb{R}^2:|z_1 - z_1^0| + |z_2 - z_2^0| \le d \right\},\notag \\ 
    &V(z_1,z_2)_{rhombus} =\chi_R(z_1,z_2),
\label{eq:rhombus_bump}
\end{align}
where $d$ is the distance between opposite vertices of the rhombus which we set to $d=3$. The shapes simulated with $V(z_1,z_2)_{circle},V(z_1,z_2)_{square},V(z_1,z_2)_{ellipse}$ and $V(z_1,z_2)_{rhombus}$ are shown in Figure \ref{fig:diff_shapes}.
 
\begin{figure}[H]
    \centering    \includegraphics[width=0.6\linewidth]{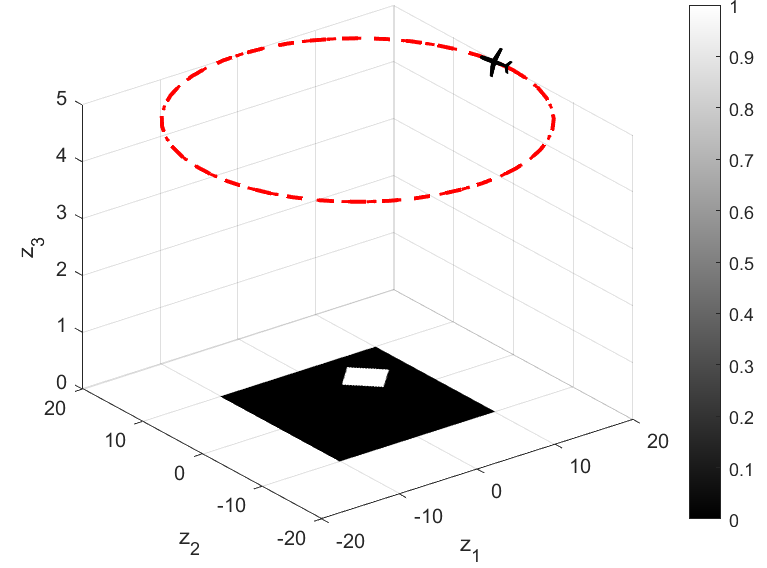}
    \caption{SAR circular flight track of radius 20 carrying an antenna located at $\gamma(s)$, for $s\in[s_{min}, s_{max}]$ flying over a terrain contained in the ROI $\Omega$ and having a rhombus shaped bump. The magnitude of the surface plot corresponds to the magnitude of $V(z_1,z_2)_{rhombus}$. The imaging antenna moves at a constant height of $h=5$ while scattering occurs at $h=0$.}
    \label{fig:Circular_Flight_Track_image}
\end{figure}

\noindent To gather SAR data, we use an antenna flying in a circular track, parametrised by $ \gamma(s)$, with $s\in [s_{min}, s_{max}]$, at a constant height $h$ above the flat terrain, as shown in Figure \ref{fig:Circular_Flight_Track_image}. The radius of the circular flight track is set to $20$ to cover the whole ROI $\Omega$. We discretise the flight path $\gamma(s)$ into 100 points as well, giving us 100 different antenna locations in the circular flight track as in \eqref{eq:flight_track_x(S)} below 
\begin{equation}
    \gamma(s) = 20\begin{bmatrix} 
\cos(\theta_s)\\ 
\sin(\theta_s) \\ 
h
\end{bmatrix}
,\qquad \text{for } \quad\theta_s \in[0,2\pi], \quad s=\{1,2,...,100\}.
\label{eq:flight_track_x(S)}
\end{equation}

\noindent We parameterise fast time $t$ as
\begin{equation}
    t=(t_i)_{i=1}^{i=100}, \qquad t_i \in [5,23], \quad \text{for} \quad i=\{1,2,...,100\}.
\end{equation}

\noindent The interval is selected to align with the backprojection algorithm (see Remark \ref{rm:interval}). Due to the parameterisations in $t,s$, we have 
\begin{equation}\label{data}
data(t,s) \in \mathbb{R}^{100\times 100}.
\end{equation}
Using $data(t,s)$ in \eqref{data}, we reconstruct the ROI $\Omega$ (or the \textit{scene}) via the backprojection algorithm \cite{nolan-cheney}. The pseudocode for the backprojection algorithm is shown in Algorithm \ref{alg:backprojection} below. 

\begin{figure}[H]
    \centering
    \includegraphics[width=\linewidth]{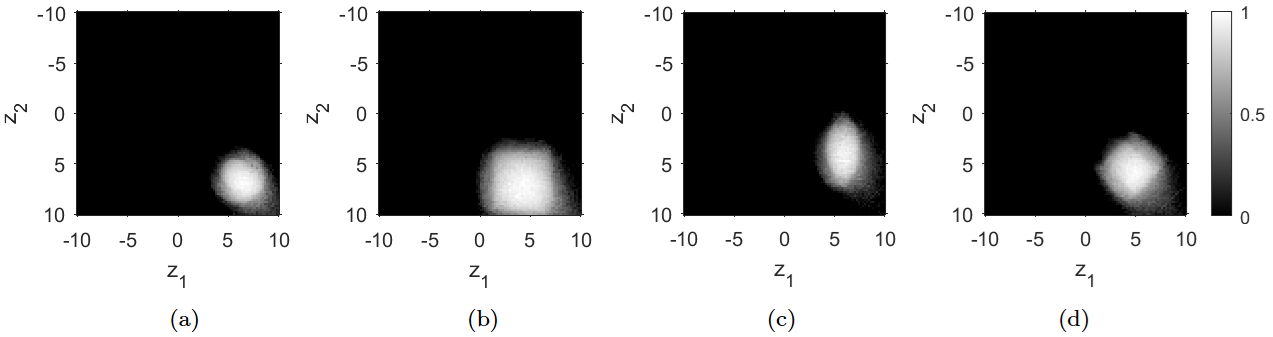}
    \caption{Backprojected reconstruction of the terrains in $\Omega$ shown in Figure \ref{fig:diff_shapes}. Data is collected from the antenna at $h=5$. The grainy and speckled images are common for SAR images. The reconstruction accurately locates the bumps. We observe some blurriness at edges which is due in part to the fact that our simplified model does not account for necessary amplitude corrections \cite{nolan-cheney}.}
    \label{fig:backprojection}
\end{figure}

\noindent The SAR backprojection algorithm (Algorithm \ref{alg:backprojection}) reconstructs an image by assigning contributions from received RADAR signals to individual image pixels. The algorithm determines the two-way travel time for each pixel by measuring the time it takes for a signal to travel from the RADAR sensor's position, $\gamma(s)$, to the pixel and back. Let us consider a sensor located at $\gamma(s)$ and measure the reflected waves at fast time $t$. If the compute two-way travel time between $\gamma(s)$ and $\Omega$ and the pixel falls within a tolerance range of $\pm tol$, the corresponding smoothed data value, $\hat{D}(t,s)$, is added to the pixel's accumulated value. Here we use $tol=0.1$. 
\\
\noindent The outcome of our reconstruction of the different shapes using backprojection (Algorithm \ref{alg:backprojection}) displayed in Figure \ref{fig:diff_shapes} is given in Figure \ref{fig:backprojection}. The backprojected image is stored using a $100\times100$ matrix, where each matrix entry corresponds to a pixel. 
\noindent The shape detection mapping with raw SAR data \eqref{eq:smooth} is
\begin{equation}
    \text{Raw SAR Data} \quad\xrightarrow[]{\text{CNN}} \quad\text{Bump Shape}
\end{equation}
with each data sample of type
\begin{equation}\label{data sample}
\textnormal{(Raw\: SAR\: Data,\: Bump\: Shape)}. 
\end{equation}
\noindent The shape detection mapping with backprojected SAR image data \eqref{eq:SAR_Image} is
\begin{equation}
    \text{Reconstructed SAR Image} \quad\xrightarrow[]{\text{CNN}} \quad\text{Bump Shape},
\end{equation}
with each data sample for classification using our CNNs being of type
\begin{equation}\label{data sample}
\textnormal{(Reconstructed\: SAR\: Image,\: Bump\: Shape)}. 
\end{equation}

\noindent To evaluate whether classification performance improves when using raw SAR data representations compared to backprojected SAR image data, we build upon the work presented in \cite{K-T-B-B}. Here, we use the raw SAR data for shape detection and compare the results with shape detection done with SAR image data. 
Therefore, our approach is summarised by the following three steps:
\begin{enumerate}
    \item Train the CNN (Section \ref{sec:cnn}, \eqref{eq:CNN_Covs} - \eqref{eq:f7}) using raw SAR data \eqref{eq:smooth} collected from antennas at heights $h\in \{0,5,10\}$.
    \item Train the CNN on backprojected SAR image data \eqref{eq:SAR_Image}, collected from antennas at heights $h\in \{0,5,10\}$.
    \item Compare the accuracy of the CNN on raw SAR data and backprojected SAR image data for each height $h\in \{0,5,10\}$.
\end{enumerate}


\begin{algorithm}[H]
\caption{Backprojection Reconstruction for SAR Imaging}
\label{alg:backprojection}
\KwIn{data, locations, t}
\KwOut{image}

Define grid: $z_1,z_2\in[-10,10]$ \;

$image \gets$ \texttt{zeros(100,100)} \;

$count \gets$ \texttt{zeros(100,100)} \;

$tolerance \gets 0.1$ 

$N \gets 100$

\For{$antenna\_position = 1$ to $N$}{
    $current\_position \gets \mathbf{x}$(antenna\_position) \;
    
    \For{$i = 1$ to $N$}{
        \For{$j = 1$ to $N$}{
            $distances(i,j) \gets\texttt{pdist2}([z_1(i), z_2(j)],current\_position) $ compute the Euclidean distance between each pixel and $current\_position$ \;
        }
    }

    \For{$t = 1$ to $N$}{
        $current\_radius \gets \sqrt{\left(\frac{c_0t}{2}\right)^2 - h^2}$  \;
        
        $id \gets$ location of grid points that are $current\_radius \pm tolerance$ from $current\_position$ 
        
        $image(id) \gets image(id) + data(t, antenna\_position)$ \;

        $count(id) \gets count(id) + 1$; \% Number of times a pixel's value is added to the image
    }
}

$image \gets image./count$ \%to ensure more often seen pixels don't get higher intensity

$image \gets$ rescale(image) \;

\KwRet image \;
\end{algorithm}

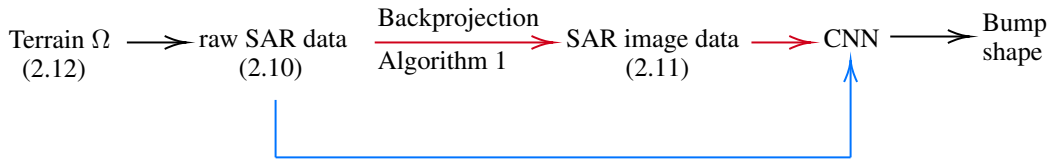
\begin{figure}[H]
\centering

\tikzset{every picture/.style={line width=0.75pt}} 

\begin{tikzpicture}[x=0.75pt,y=0.75pt,yscale=-1,xscale=1]

\draw    (65,66) -- (95,66) ;
\draw [shift={(95,66)}, rotate = 180] [color={rgb, 255:red, 0; green, 0; blue, 0 }  ][line width=0.75]    (10.93,-3.29) .. controls (6.95,-1.4) and (3.31,-0.3) .. (0,0) .. controls (3.31,0.3) and (6.95,1.4) .. (10.93,3.29)   ;
\draw [color={rgb, 255:red, 208; green, 2; blue, 27 }  ,draw opacity=1 ]   (190,66) -- (280,66) ;
\draw [shift={(280,66)}, rotate = 180] [color={rgb, 255:red, 208; green, 2; blue, 27 }  ,draw opacity=1 ][line width=0.75]    (10.93,-3.29) .. controls (6.95,-1.4) and (3.31,-0.3) .. (0,0) .. controls (3.31,0.3) and (6.95,1.4) .. (10.93,3.29)   ;
\draw [color={rgb, 255:red, 208; green, 2; blue, 27 }  ,draw opacity=1 ]   (380,66) -- (410,66) ;
\draw [shift={(410,66)}, rotate = 180] [color={rgb, 255:red, 208; green, 2; blue, 27 }  ,draw opacity=1 ][line width=0.75]    (10.93,-3.29) .. controls (6.95,-1.4) and (3.31,-0.3) .. (0,0) .. controls (3.31,0.3) and (6.95,1.4) .. (10.93,3.29)   ;
\draw    (450,63) -- (490,63) ;
\draw [shift={(490,63)}, rotate = 180] [color={rgb, 255:red, 0; green, 0; blue, 0 }  ][line width=0.75]    (10.93,-3.29) .. controls (6.95,-1.4) and (3.31,-0.3) .. (0,0) .. controls (3.31,0.3) and (6.95,1.4) .. (10.93,3.29)   ;
\draw [color={rgb, 255:red, 0; green, 116; blue, 255 }  ,draw opacity=1 ]   (140,95) -- (140,124) ;
\draw [color={rgb, 255:red, 0; green, 116; blue, 255 }  ,draw opacity=1 ]   (140,124) -- (430,124) ;
\draw [color={rgb, 255:red, 0; green, 116; blue, 255 }  ,draw opacity=1 ][fill={rgb, 255:red, 0; green, 0; blue, 0 }  ,fill opacity=1 ]   (430,124) -- (430,75) ;
\draw [shift={(430,75)}, rotate = 90] [color={rgb, 255:red, 0; green, 116; blue, 255 }  ,draw opacity=1 ][line width=0.75]    (10.93,-3.29) .. controls (6.95,-1.4) and (3.31,-0.3) .. (0,0) .. controls (3.31,0.3) and (6.95,1.4) .. (10.93,3.29)   ;

\draw (4,58) node [anchor=north west][inner sep=0.75pt]   [align=left] {Terrain $\Omega$ \\ \ \ \eqref{Omega}};
\draw (100,58) node [anchor=north west][inner sep=0.75pt]   [align=left] {raw SAR data \\  \ \ \ \ \ \ \eqref{eq:smooth}};
\draw (285,58) node [anchor=north west][inner sep=0.75pt]   [align=left] {SAR image data \\     \ \ \ \ \ \ \ \ \ \eqref{eq:SAR_Image}};
\draw (415,58) node [anchor=north west][inner sep=0.75pt]   [align=left] {CNN};
\draw (495,50) node [anchor=north west][inner sep=0.75pt]   [align=left] {Bump\\shape};
\draw (190,69) node [anchor=north west][inner sep=0.75pt]   [align=left] {Algorithm \ref{alg:backprojection}};
\draw (190,47) node [anchor=north west][inner sep=0.75pt]   [align=left] {Backprojection};
\end{tikzpicture}
\caption{Schematic description of the process followed for our shape detection classification task using SAR data. In case 1 (top line represented via red arrows), the raw SAR data is backprojected to form an image (Algorithm \ref{alg:backprojection}) fed to our CNN (Section \ref{sec:cnn}) In case 2 (bottom line represented via blue arrow), the raw SAR data is fed to the CNN directly. } 

\label{fig:2ShapeComparison}
\end{figure}

\vspace{1em}
This allows us to determine which approach - using raw SAR data or image-based SAR data - leads to better shape classification performance, extending the findings presented in \cite{K-T-B-B}. A visual comparison of these two approaches is shown in Figure \ref{fig:2ShapeComparison} and results of classification for both approaches is shown in Figure \ref{fig:ShapeComparison}.

\noindent We simulate 4000 data points, with 1000 points for each shape of the bump (i.e., 1000 circles, 1000 squares, 1000 ellipses, 1000 rhombuses). The centre of the bump, $(z_1^0, z_2^0)\in\Omega$, varies for each sample and is arbitrarily chosen as per \eqref{eq:centre}. This ensures that the CNN can detect the shape, regardless of its position on the scene grid $\Omega$. We use 80\% of the data (3200 samples) for training. The remaining data is split as 10\%  for validation (400 samples) and 10\% for testing (400 samples). The breakdown of the data into training, validation and testing is a well-known aspect of machine learning, which is further explained in Section \ref{sec:shape_results}.


\subsection{Data simulation for multi-scatterer detection}
As we establish in Section \ref{sec:shape_results} that raw SAR data provides superior classification than backprojected SAR images, we use raw SAR data only for the multi-scatterer detection treated in Section \ref{sec:multi_Scatter}. The ROI $\Omega$ for multi-scatterer detection is that defined in \eqref{Omega}. The antenna collecting the raw SAR data is placed at a height of $h=5$, flying in a circular flight track as in \eqref{eq:flight_track_x(S)}. We use a height $h=5$ here as this is the height achieving the highest classification accuracy in Section \ref{sec:shape_results}. The primary objective of multi-scatterer detection is to identify the maximum and minimum radii of scatterers that the CNN introduced in Section \ref{sec:cnn}, \eqref{eq:CNN_Covs} - \eqref{eq:f7} can accurately detect without confusing multiple scatterers for a single large bump. The pseudocode outlining this process is shown in Algorithm \ref{alg:mult_scat}. This gives an indication regarding the resolution limit of our classification. To achieve this, we'll train our CNN to classify terrains based on the number of scatterers present. The CNN's specific task is to determine whether a terrain within the ROI $\Omega$ contains one or two scatterers using raw SAR data. We progressively increase the radii of these scatterers, training a new CNN for each radius $r \in \{1,2,3,4,5,10,15\}$ until we identify the minimum and maximum radii ($r_{max}$ and $r_{min}$ respectively) that yield 100.00\% classification accuracy. With the steps outlined in Algorithm \ref{alg:mult_scat}, we effectively determine the range of scatterer radii that are needed for 100.00\% detection of multi-scatterers.

\vspace{1em}
\begin{algorithm}[H]
\caption{Multi-Scatterer detection algorithm}
\label{alg:mult_scat}
\KwIn{data, locations, t}
\KwOut{$[r_{min},r_{max}]$}

Define grid: $\Omega$ 

$r \in \{1, 2, 3,4,5, 10, 15\}$

\For{each $r$}{
 Generate 2500 samples with one circular bump of radius = $r$ placed arbitrarily in $\Omega$.

 Generate 2500 samples with two circular bumps of radius = $r$ placed arbitrarily in $\Omega$.

Split the generated 5000 samples into 80\% training (4000 Samples), 10\% testing(250 Samples) and 10\% validation data (250 Samples)

Train CNN (Section \ref{sec:cnn}) to detect whether terrains contain one or two circular bumps \eqref{eq:circular_bump}.

Determine the accuracy of multi-scatterer detection on Test data.

}

$r_{min} \gets$ Minimum $r$ that gives 100\% accuracy in detection.

$r_{max} \gets$ Maximum $r$ that gives 100\% accuracy in detection.

\KwRet [$r_{min},r_{max}$] 
\end{algorithm}
\vspace{1em}


\subsection{Number of scatterer detection}
To detect the exact number of scatterers present in a terrain, we generate 6000 terrains. The terrains are as defined in \eqref{Omega}. Each terrain contains either 1, 2, or 3 scatterers. The scatterers are circular in nature as defined by \eqref{eq:circular_bump} of radius, $r=2$. We use $r=2$ because it was empirically determined to be the smallest radius of a scatterer that can be detected with complete accuracy by a CNN (Section \ref{sec:multi_Scatter}). We use raw SAR data to detect the number of scatterers using CNNs. For the reason mentioned above, the raw SAR data is collected by an antenna placed at a height of $h=5$, flying on a circular path (Equation~\eqref{eq:flight_track_x(S)}). Furthermore, we draw a comparison of the results of number of scatterers detected to number of inclusions detection with EIT in Section \ref{sec:EITComp}.

\subsection{Ice-type detection}
In the final experiment, we detect different types of ice in Greenland using SAR imagery \cite{data}. The dataset contains SAR images collected from the Sentinel-1 satellite over the Belgica Bank in Greenland, which have been classified by a subject matter expert into one of the eight types listed in Table \ref{tb:Ice_Type}.
\begin{remark}
     The dataset in \cite{data} contains SAR images processed after data collection from satellite Sentinel-1; therefore, we do not perform any backprojection/simulation as the data is already formatted to be usable for ML. This dataset contains SAR images only, therefore we do not use raw SAR data but the SAR image data for classification. 
\end{remark}
\noindent The dataset in \cite{data} contains processed SAR images of different regions of ice. Each image is a pixel in a grid of $256\times 256$. Each image has a corresponding ice-type label associated with it as shown in Table \ref{tb:Ice_Type}.

\begin{table}[H]
\centering
\begin{tabular}{lc}
\textbf{Ice Type} & \textbf{Label O} \\ \hline
Black Border      & 1                \\
Old Ice           & 2                \\
First Year Ice    & 3                \\
Glaciers          & 4                \\
Icebergs          & 5                \\
Mountains         & 6                \\
Young Ice         & 7                \\
Water group       & 8               
\end{tabular}
\caption{Different types of ice as detected in SAR images \cite{data} and their corresponding CNN class labels.}
\label{tb:Ice_Type}
\end{table}

\noindent In the case of ice-type detection, we do not use all of the data available in the dataset \cite{data}. Instead, we prioritise the use of a balanced representation of classes to ensure that the CNN learns to classify different types of ice without the risk of overfitting. We refer to \cite{Bu-Ma-Ma} for further reading on this matter. Specifically, we use 9360 samples with each class label equally represented. As before, we use 80\% of the data for training (7488 samples), the remaining 10\% for validation (936 samples) and the last 10\% for testing (936 samples). The mapping for ice type detection is
\begin{equation}
    \text{Sentinel-1 SAR Image}\quad \xrightarrow[]{\text{CNN}} \quad\text{Ice \:Type}.
\end{equation}
Each sample for ice-type detection is of type
\begin{equation}
\textnormal{(SAR\: Image, \: Ice \: Type)}.
\end{equation}

 \noindent The results of ice-type detection with SAR and CNNs are detailed in Section \ref{sec:ice_results}. 


\section{Results}\label{sec:results}
In this section, we present the classification results of various SAR datasets using the CNN introduced in Section \ref{sec:cnn}, \eqref{eq:CNN_Covs} - \eqref{eq:f7}.


\subsection{Shape detection}
\label{sec:shape_results}
We evaluate the CNN's performance in classifying SAR data into different shapes. Following the approach of \cite{K-T-B-B}, we compare two input types: raw SAR data and backprojected images derived from the raw SAR data. Our results show that raw SAR data achieves higher accuracy in shape detection. 

\begin{assumption}
    There is one bump present in the terrain within $\Omega$ \eqref{Omega}. The bump can be of one of four shapes defined in Table \ref{tb:shape_types}.
\end{assumption}

We utilise a dataset consisting of 4000 samples, which are split as 80\% (3200 samples) for training, 10\% (400 samples) for validation, and 10\% (400 samples) for testing. The validation dataset helps estimate the model's performance during training while adjusting weights and biases as per \eqref{eq:CNN:min}. The test dataset, withheld during training, provides an unbiased evaluation of the final model's performance \eqref{eq:CNN:min}-\eqref{eq:cross}. We start with an antenna at height $h=10$, by then decreasing the hight to $h=5$ and $h=0$.

\begin{remark}\label{rm:interval}
    There is an inherent trade-off between the sensor altitude and the radius of the flight path. Additionally, in SAR, image resolution is not determined solely by the platform’s altitude. Instead, it heavily depends on the diversity of angular perspectives collected during the RADAR’s motion. These perspectives govern the spatial frequency content of the scene, which directly impacts resolution, especially through the amplitude and phase of the collected signals. However, as a proof-of-concept, we use altitude-only measurements in this paper.

\end{remark}

\begin{figure}[H]
    \centering
    \includegraphics[width=0.8\linewidth]{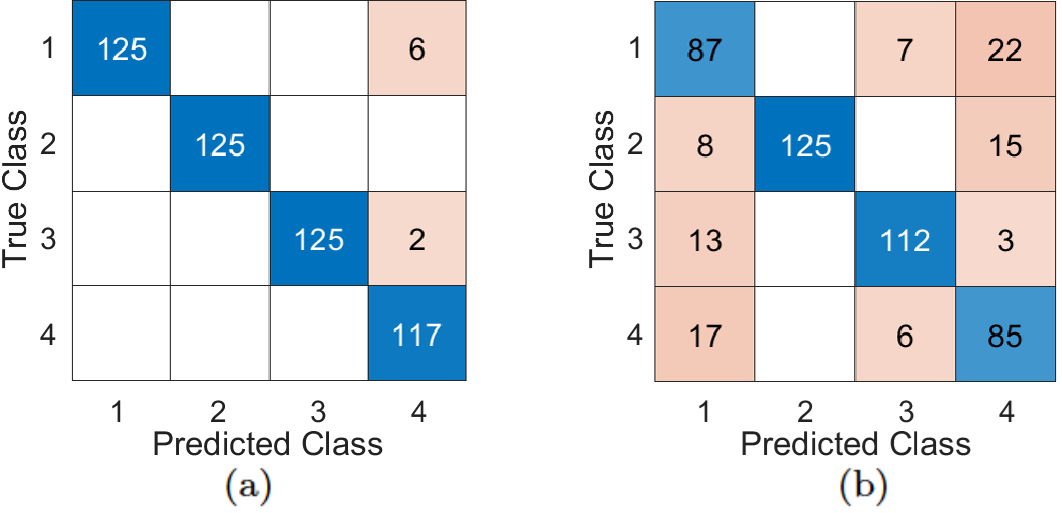}
    \caption{Figure \ref{fig:h10Shape}(a) displays the confusion matrix for shape detection using raw SAR data \eqref{eq:smooth}, collected from an antenna at a height of $h=10$ showing 98.40\% accuracy. In contrast, Figure \ref{fig:h10Shape}(b) illustrates the shape detection results obtained from backprojected SAR image data as per \eqref{eq:SAR_Image}, also gathered from an antenna at $h=10$ showing 81.80\% accuracy.}
    \label{fig:h10Shape}
\end{figure}

\begin{figure}[H]
    \centering
    \includegraphics[width=0.8\linewidth]{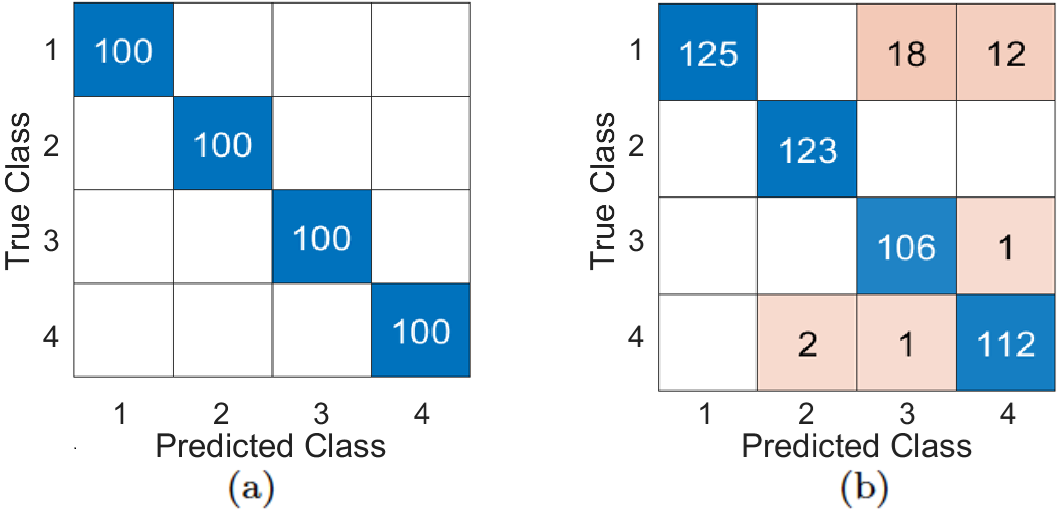}
    \caption{Figure \ref{fig:h5Shape}(a) displays the confusion matrix for shape detection using raw SAR data \eqref{eq:smooth}, collected from an antenna at a height of $h=5$ showing 100.00\% accuracy. Figure \ref{fig:h5Shape}(b) illustrates the shape detection results obtained from backprojected SAR image data as per \eqref{eq:SAR_Image}, also gathered from an antenna at $h=5$ showing 93.20\% accuracy.}
    \label{fig:h5Shape}
\end{figure}

\begin{figure}[H]
    \centering
    \includegraphics[width=0.8\linewidth]{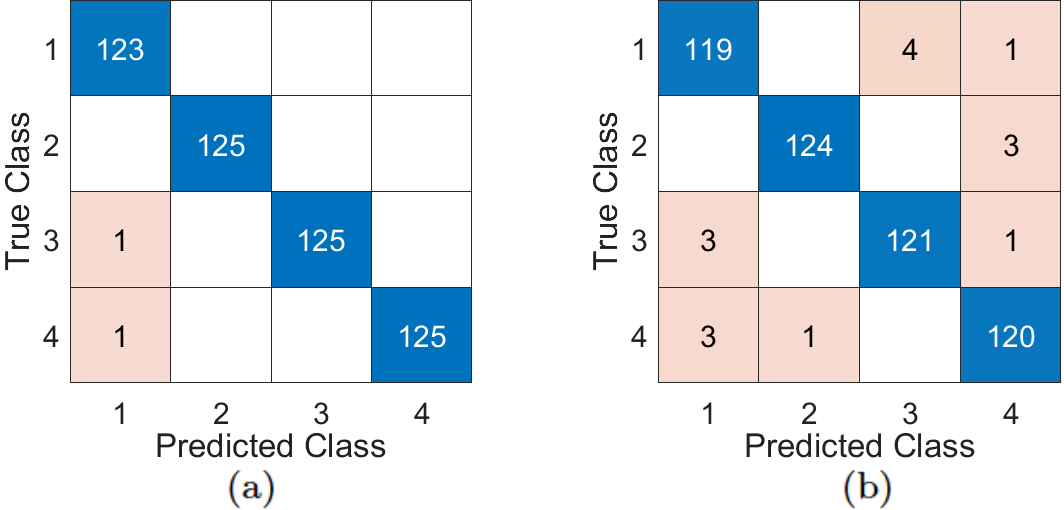}
    \caption{Figure \ref{fig:h0Shape}(a) displays the confusion matrix for shape detection using raw SAR data \eqref{eq:smooth}, collected from an antenna at a height of $h=0$ showing 99.60\% accuracy. Figure \ref{fig:h0Shape}(b) shows the shape detection results obtained from backprojected SAR image data as per \eqref{eq:SAR_Image}, also gathered from an antenna at $h=0$ showing 96.80\% accuracy.}
    \label{fig:h0Shape}
\end{figure}

\noindent When the antenna flies at the fixed hight $h=5$ and $h=0$, respectively, the classification results for shape detection via the CNN of Section \ref{sec:cnn} on raw SAR data \eqref{eq:smooth} and backprojected SAR image data \eqref{eq:SAR_Image} are presented in Figures \ref{fig:h5Shape} and \ref{fig:h0Shape}, respectively.

\noindent From Figures \ref{fig:h10Shape}, \ref{fig:h5Shape} and \ref{fig:h0Shape} we see that raw SAR data \eqref{eq:smooth} outperforms backprojected SAR image data \eqref{eq:SAR_Image} in classification of shapes via our CNN of Section \ref{sec:cnn}. Here, the case of $h=0$ is particularly interesting, as it can be compared to a temporally static model with a 2D cross-section, similar to Electrical Impedance Tomography (EIT) \cite{Bo,A-G-L}. We are particularly interested in comparing this case with the results obtained in \cite{G-H-N-N}, where neural networks were not used to classify the precise shape of inclusions; instead, the study focused on detecting their size, number, as well as the presence of anisotropy within the inclusions.
We compare our results with those of \cite{G-H-N-N}, noting a key distinction: while \cite{G-H-N-N} employed a static EIT model where the Dirichlet-to-Neumann matrix was fed into an artificial neural network (ANN), the current approach utilises a SAR dynamic model to image a terrain containing an inclusion. Using EIT and ANNs, a high accuracy ($\geq90\%$) was achieved in classification tasks. Here, we achieve an accuracy of $100\%$ when using raw SAR data that are collected at height $h=5$. 
\begin{remark}
    There is a fundamental difference in the nature of the tasks of EIT and SAR classifications. In \cite{G-H-N-N}, the variations in input were driven by differences in the conductivity $\sigma$ of the material under investigation, allowing $\sigma$ to be anisotropic, i.e. represented by a matrix-valued function.  In contrast, in the present work, the input differences arise from the scalar-valued reflectivity function, $ V(z_1,z_2)$. 
\end{remark}

\begin{figure}[H]
    \centering
    \includegraphics[width=0.5\linewidth]{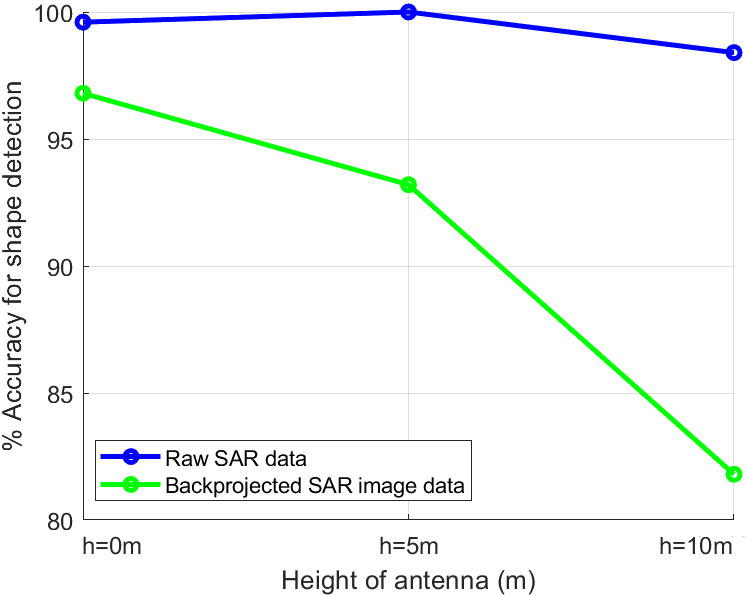}
    \caption{Comparison of shape detection classifications for SAR data. The blue line represents the classification accuracy using raw SAR data \eqref{eq:smooth}, while the green line shows the accuracy achieved with SAR image data \eqref{eq:SAR_Image}. The raw SAR data and backprojected SAR image data are collected at three different antenna heights $h \in \{ 0,5,10\}$. Raw SAR data (blue line) consistently has higher accuracy in shape detection than backprojected SAR image data (green line).} 
    \label{fig:ShapeComparison}
\end{figure}

\noindent The performance accuracies of our CNN (Definition \ref{def:acc}) trained on raw SAR data \eqref{eq:smooth} and backprojected SAR images \eqref{eq:SAR_Image} for shape detection is shown in Figure \ref{fig:ShapeComparison}. We see that across all tested antenna heights ($h \in \{0,5,10\}$), raw SAR data \eqref{eq:smooth} outperforms backprojected SAR image data \eqref{eq:SAR_Image} in shape detection. This reinforces the advantage of using raw SAR data for tasks like shape detection, as it retains more information than backprojected images, even at improved resolutions. 

\noindent As shown in Figure \ref{fig:ShapeComparison}, raw SAR data \eqref{eq:smooth} consistently outperforms backprojected SAR image data \eqref{eq:SAR_Image} in classification accuracy. We also observe that lowering the antenna improves classification performance on SAR image data, likely due to the higher angular resolution of the reconstructed images. However, even at the lowest antenna height $h=0$, the classification accuracy of our CNN on backprojected SAR image data \eqref{eq:SAR_Image} does not match that achieved with raw SAR data \eqref{eq:smooth}. This further underscores the advantage of using raw SAR data for shape and object detection tasks, as it preserves crucial information lost during the backprojection process (Algorithm \ref{alg:backprojection}). Our findings, extend the conclusions of \cite{K-T-B-B} that classification using ML is superior on raw SAR data over SAR image data.


\subsection{Multi-Scatterer Detection}\label{sec:multi_Scatter}
In this section, we aim to determine the maximum size multiple scatterers located in our ROI $\Omega$ \eqref{Omega} can reach before the CNN \eqref{f1}-\eqref{eq:f7} begins to misclassify them as a single large bump. As a side exercise, we also analyse what the minimum size of the bumps should be to ensure perfect accuracy of our CNN classification.  We design the experiment to gain insight into the resolution of the SAR system in order to have a reliable classification. We consider the simple case of multiple circular bump scatterers \eqref{eq:circular_bump}, where the task is to determine the maximum radii scatterers are allowed to have in order for our CNN to classify them as distinct objects. Given the superior performance of shape detection on raw SAR data \eqref{eq:smooth} compared to backprojected image data \eqref{eq:SAR_Image}, we utilise raw SAR data only for this task. As Section \ref{sec:shape_results} demonstrated that using raw SAR data collected from an antenna flying along the flight path \eqref{eq:flight_track_x(S)} at constant height $h=5$ yields the highest classification accuracy (100.00\%), we proceed the multi-scatterer detection task by collecting raw SAR data for our CNN with an antenna positioned at this height. To determine the largest bump size detectable by our CNN using raw SAR data, we place the bumps in close proximity. This allows us to observe if the CNN can differentiate between two large, overlapping bumps. To do this, we choose the centre of the first bump to be in the range

\begin{equation}
    0 \leq z_1^0,z_2^0 \leq 5,
\end{equation}
and if there is a second bump present, we choose the center of the second bump as
\begin{equation}
    -1 \leq \tilde{z}_1^0,\tilde{z}_2^0 \leq -4.
\end{equation}

\begin{assumption}
    There is at least one circular bump \eqref{eq:circular_bump} in the terrain $\Omega$ \eqref{Omega}. The terrain can contain at most two circular bumps. 
\end{assumption}

\begin{figure}[H]
    \centering
    \includegraphics[width=1\linewidth]{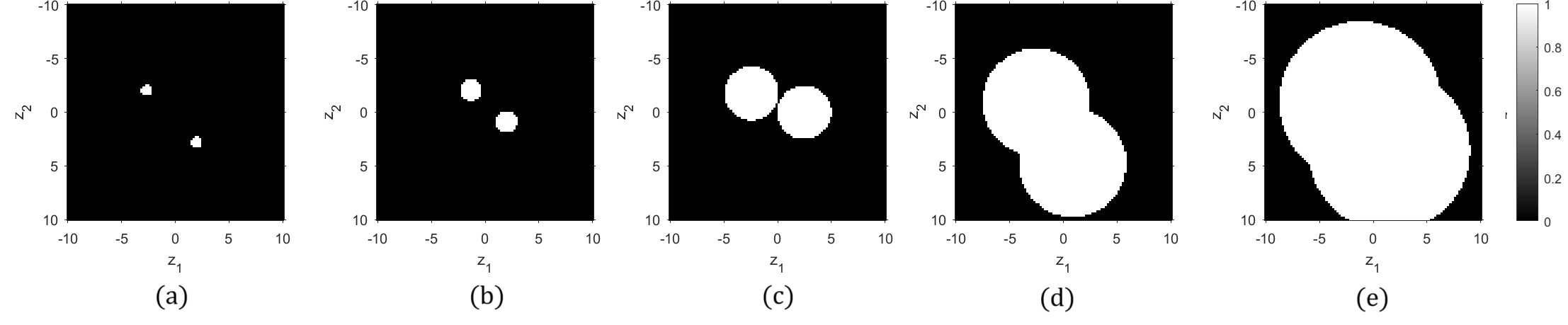}
    \caption{Terrains within our ROI $\Omega$ containing multi-scatterers of varying radii. Figures \ref{fig:MultBump}(a) - (e) show, respectively, circular scatterers of radii $r \in \{ 1,2,5,10,15$\}.}
    \label{fig:MultBump}
\end{figure}

We start by testing whether the raw SAR data \eqref{eq:smooth} can accurately distinguish between two small bumps. We then gradually increase the size of the bumps and repeat the classification process. The radius $r$ of the circular bump is one of $r = 1,2,3,4,5,10,15$. Figure \ref{fig:MultBump} shows the different-sized bumps used in this experiment. 

\noindent We continue the classification process until we observe a decline in classification accuracy, allowing us to identify the smallest and largest bump size that can still be reliably detected by our CNN with the raw SAR data (see Algorithm \ref{alg:mult_scat}). The results of multi-scatterer detection are given in Table \ref{tb:mult_scat_tb}.

\begin{table}[H]
\centering
\begin{tabular}{cr}
\textbf{Scatterer radius r} & \textbf{Accuracy} \\ \hline
1                         & 98.25\%           \\
2                         & 100.00\%           \\
3 & 100.00\% \\
4 & 100.00\%\\
5                        & 92.75\%          \\
10                        & 91.00\%          \\
15 & 84.00\%\\
\end{tabular}
\caption{Performance accuracy of our CNN with raw SAR data on detection of multiple scatterers of varying radii. We see that we obtain a 100.00\% accuracy in multi-scatterer detection for $r =2,3,4$.}
\label{tb:mult_scat_tb}
\end{table}

\noindent We observe that, in this experiment, 100\%-accuracy was achieved for $2 \leq r \leq 4$, or
\begin{equation}
\begin{aligned}
    1.58\% - 25.14\% \text{ of the area of ROI }\Omega.
\end{aligned}
    \label{eq:r_range}
\end{equation}

As expected from an intuitive point of view, if the two bumps occupy a majority of the ROI $\Omega$ region, achieving $100\%$ accuracy in multi-scatter detection might not be feasible, although the performance level remains quite good to allow to distinguish the main features of the circular bumps in Figures \ref{fig:MultBump}(d) and (e) where the radii are $r=10$ and $r=15$, respectively.


\subsection{Comparison with Electrical Impedance Tomography (EIT)}
\label{sec:EITComp}
In this section we consider again the scenario where the terrain within our ROI $\Omega$ \eqref{Omega} has one circular bump only and the task for our CNN is to detect the radius of the single scatterer using raw SAR data \eqref{eq:smooth} only. The antenna is flying at a fixed hight $h=0$, allowing us to compare the results of this experiment with the EIT findings in \cite{G-H-N-N}. A phantom within a circular tank was considered in \cite{G-H-N-N} and electrostatic measurements performed on the boundary of the tank. The task was, as in the current Section, to classify the radius of an inclusion within the tank having four different sizes as here.

\begin{assumption}
    There is one circular bump \eqref{eq:circular_bump} in the terrain within our ROI $\Omega$ \eqref{Omega}. The circular bump can be of one of four radii in \eqref{eq:sigma}.
\end{assumption}

Here, the CNN in Section \ref{sec:cnn} is tasked with the following experiment, summarised by the following schematic map:
\begin{equation}
\text{Raw SAR data with reflectivity parameter $r$} \quad \xrightarrow[]{\text{CNN}}\quad
    \begin{cases}
    r = 1 & \text{Class 1} \\
    r = 2 & \text{Class 2} \\
    r = 5 & \text{Class 3} \\
    r = 10 & \text{Class 4}
    \end{cases}.
    \label{eq:sigma}
\end{equation}

\begin{figure}[H]
     \centering
     \includegraphics[width=0.4\linewidth]{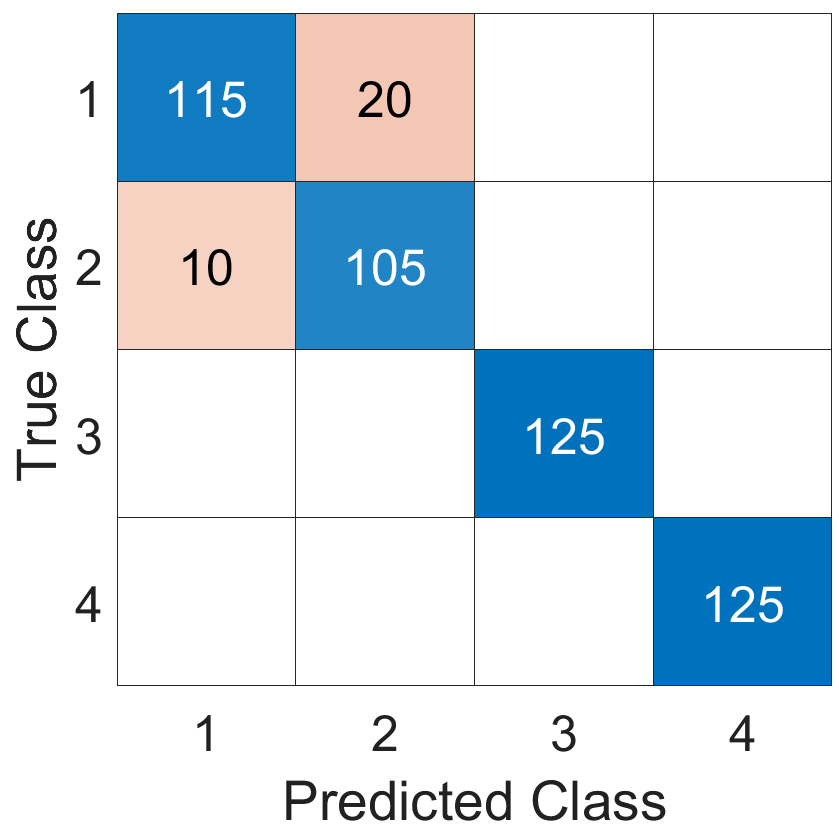}
     \caption{Confusion Matrix for bump radii detection for the classification task given in \eqref{eq:sigma} using raw SAR data collected by an antenna flying at a fixed height $h=0$ from terrains within our ROI $\Omega$ \eqref{Omega}. The ROI contains a circular scatterer only, having radius one of the values given in \eqref{eq:sigma}. The confusion matrix shows a 94.00\% accuracy.}
     \label{fig:ScatterRadii}
 \end{figure}

\noindent In this experiment, we generate 5000 samples, with 1250 samples in each class given in \eqref{eq:sigma}. We use 80\% of the data for training (4000 samples), 10\% of the remaining data (500 samples) for validation and the last 10\% for testing (500 samples). The Confusion Matrix of the training data is shown in Figure \ref{fig:ScatterRadii}.

\noindent We are able to detect the radius of a single scatterer in this classification task with 94.00\% accuracy. In this regard, the static EIT imaging modality treated in \cite{G-H-N-N} performs better, with 100\% accuracy. On the other hand, EIT failed (31.30\% accuracy) in detecting the number of inclusions (bumps) in the tank in \cite{G-H-N-N}. To investigate whether SAR outperforms EIT in this regard, we simulate terrains containing either 1,2 or 3 scatterers (bumps). 
 \begin{assumption}
     The terrains within our ROI $\Omega$ \eqref{Omega} contain either 1,2 or 3 circular bumps \eqref{eq:circular_bump} of a fixed radius $r=2$.
 \end{assumption}
 
\begin{figure}[H]
     \centering
     \includegraphics[width=0.35\linewidth]{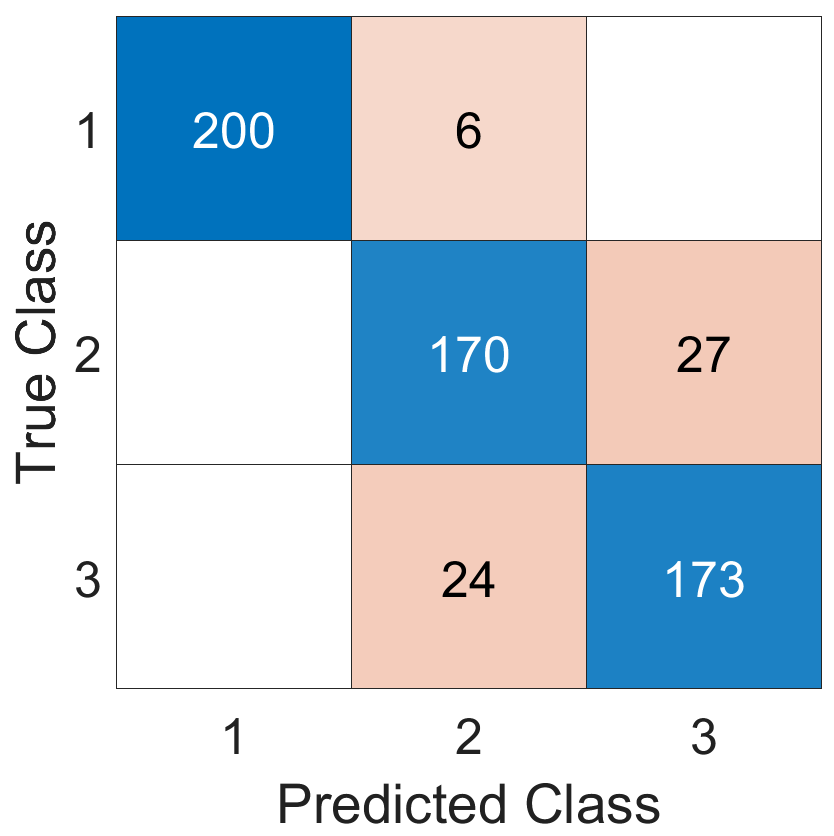}
     \caption{Confusion Matrix for detecting the number of scatterers of a fixed radius $r=2$ using raw SAR data collected from an antenna flying on the circular path of Figure \ref{fig:Circular_Flight_Track_image} at a fixed height $h=0$, showing 90.50\% accuracy.}
     \label{fig:IncNum}
 \end{figure}

\noindent Our CNN is tasked with classifying the terrain within $\Omega$ as containing 1,2, or 3 scatterers (bumps) as summarised by the schematic map:
 \begin{equation}
\text{Raw SAR data } \quad \xrightarrow[]{\text{CNN}}\quad
    \begin{cases}
    \text{1 Bump} & \text{Class 1} \\
    \text{2 Bumps} & \text{Class 2} \\
    \text{3 Bumps} & \text{Class 3} \\
    \end{cases}.
\end{equation}
 We generate 6000 terrains. We use 4800 for training, 600 samples for testing, and another 600 samples for validation. The results of this training are given in the Confusion Matrix in Figure \ref{fig:IncNum}.

This last experiment suggests that a dynamic imaging method such as SAR may outperform EIT \cite{G-H-N-N} when it comes to detecting multiple scatterers.


\subsection{Ice-type detection}
\label{sec:ice_results}
To validate the real-world applicability of combining SAR images and the CNN introduced in Section \ref{sec:cnn}, we test our CNN on a dataset of SAR images from the Sentinel-1 satellite \cite{data} in the Belgica Bank. This dataset consists of images that are $256 \times 256$-pixel grid. For details of the Sentinel-1 mission we refer to \cite{Sentinel1}. As a result, we need to modify the CNN architecture outlined in \eqref{eq:CNN_Covs} of Section \ref{sec:cnn} to accommodate the input size $256 \times 256$. The corresponding tensor dimensions in the CNN \eqref{f1}-\eqref{eq:f7} are given in Table \ref{tb:cnn}.

We set the number of filters in \eqref{f1} to be $K=16$ as the ice type detection problem is more complex than the problems studied in Sections \ref{sec:shape_results}-\ref{sec:EITComp} where we used $K=1$. Here we deal with more classes, more intricate patterns to differentiate and larger images as listed in Tables \ref{tb:cnn} and \ref{tb:Ice_Type}. 

\begin{remark}
    We do not have raw SAR data of the Belgica Bank in the dataset \cite{data}. Therefore, for the process of ice type detection, we use backprojected SAR images gathered from Sentinel-1. 
\end{remark}

\begin{figure}[H]
    \centering
    \includegraphics[width=0.5\linewidth]{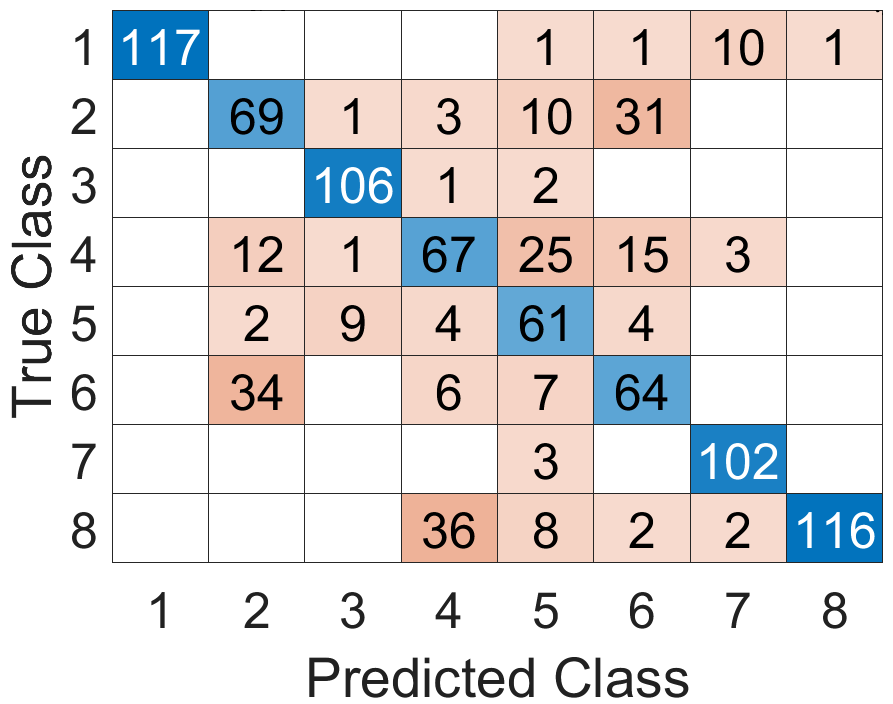}
    \caption{Confusion Matrix for ice type detection via our CNN introduced in Section \ref{sec:cnn} and  the SAR image data from the Sentinel-1 satellite \cite{data} in the Belgica Bank.}
    \label{fig:CM_Ice}
\end{figure}

\begin{figure}[H]
    \centering
    \includegraphics[width=0.8\linewidth]{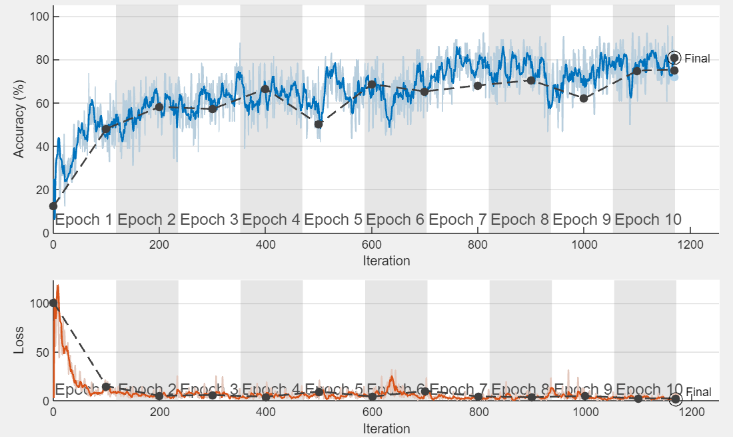}
    \caption{Training progress for our CNN with $K=16$ filters for ice type detection. The black dotted lines represent the accuracy progression on the validation dataset, the blue lines show the training accuracy, and the orange lines indicate the loss in (\ref{eq:cross}) over time.}
    \label{fig:ice_training}
\end{figure}

\noindent Here we train the CNN for 100 iterations. The training progress is shown in Figure \ref{fig:ice_training}. We see a training accuracy of 81.00\%. When tested on the test dataset of 938 samples, the trained CNN performed with an accuracy of 75.00\%. The Confusion Matrix for ice type detection is given in Figure \ref{fig:CM_Ice}. 

\section{Conclusion}
\label{sec:conclusion}
The application of ML techniques to the inverse problem of detecting objects in SAR raw data and reconstructed images is explored. Specifically, using both real and simulated datasets of raw SAR data and their associated backprojected images, four problems are addressed in this paper: i) detecting the shape of an object; ii) distinguishing multiple scatterers; iii) detecting the radius of a single circular scatterer and the number of multi-scatters, drawing a comoparision with EIT; iv) detecting the type of sea ice found in the Balgica Bank \cite{data}.
We employ the CNN \eqref{f1}-\eqref{eq:f7} for these tasks and achieve an accuracy of 100\% in detecting the shape of an object (among the different shapes listed in Table \ref{tb:shape_types}) using raw SAR data (Section \ref{sec:shape_results}). Our results confirm the importance of using raw data (as opposed to backprojected data) during SAR classification tasks \cite{K-T-B-B} using ML. 
We also detect the number of scatterers with 90.50\% accuracy and the radius of a single scatterer with 94\% accuracy, when the antenna is moving on the ground, i.e., along the circular path of Figure \ref{fig:Circular_Flight_Track_image} at the fixed hight $h=0$ (Section \ref{sec:EITComp}), while drawing a parallel comparison with the static imaging modality EIT \cite{G-H-N-N}. Finally, we achieve an accuracy of 75\% in classifying the 8 different types of ice (Table \ref{tb:Ice_Type}) using SAR images only, as raw data for this classification task was not available (Section \ref{sec:ice_results}). For this final task, we tested our CNN on a dataset of SAR images from the Sentinel-1 satellite \cite{data} in the Belgica Bank. A summary of the results conducted in this paper is given in Table \ref{tb:SAR_Results}.

\begin{table}[H]
\centering
\begin{tabular}{llcc}
\textbf{Task}               & \textbf{Data type}     & \textbf{$h$}      & \textbf{Accuracy} \\ \hline
Shape Detection             & Raw SAR Data &0           & 99.90\%            \\
Shape Detection             & SAR Image data & 0  &96.80\%            \\
Shape Detection             & Raw SAR data & 5  &100.00\%            \\
Shape Detection             & SAR Image 
data & 5  & 93.20\%            \\
Shape Detection             & Raw SAR data & 10 & 98.40\%            \\
Shape Detection             & SAR Image data & 10 & 81.80\%            \\
Radius of Scatter Detection & Raw SAR Data &5            & 94.00\%            \\

Number of Scatter Detection & Raw SAR Data  &5          & 90.50\%           \\
Ice Type Detection          & SAR Image Data   &   693km \cite{Sentinel1}   & 75.00\%          
\end{tabular}
\caption{Summary of classification results in this paper. Here $h$ is the height of the SAR antenna.}
\label{tb:SAR_Results}
\end{table}

\noindent It is anticipated that this proof-of-concept paper will serve as a first step for future work by the same authors on identifying more complex scatterers using both raw SAR data and SAR backprojected images.

\noindent A possibility for improving our results could be the use of multiple flight tracks and a more comprehensive reconstruction algorithm than that given in Algorithm \ref{alg:backprojection}. We may also enhance our single scattering model (Section \ref{sec:single_scatter}),  incorporate more realistic sources, as well as other enhancements to make our method widely applicable.

\section*{Acknowledgments}
This publication has emanated from research conducted with the financial support of Research Ireland under Grant number 18/CRT/6049. For the purpose of Open Access, the author has applied a CC BY public copyright license to any Author Accepted work version arising from this submission. The authors would like to thank the Isaac Newton Institute for Mathematical Sciences, Cambridge, for support and hospitality during the programme Rich and Nonlinear Tomography - a multidisciplinary approach, where work on this paper was undertaken. This work was supported by EPSRC grant EP/R014604/1.

\end{document}